%% file: main.tex
\documentclass{article} 
\usepackage{iclr2026_conference,times}
\usepackage{graphicx} 
\usepackage{amsmath}   
\usepackage{amsthm}    

\usepackage{rotating}
\usepackage{amssymb}
\usepackage{xcolor}
\usepackage{float}
\usepackage{caption}
\usepackage{adjustbox}
\usepackage{algorithm}
\usepackage{algorithmic}
\usepackage{appendix}
\usepackage{caption}

\usepackage[dvipsnames]{xcolor}
\usepackage{tocloft} 
\usepackage{etoolbox} 

\newlistof{appendixtoc}{apt}{Information Estimation with Discrete Diffusion Supplementary Material}
\newcounter{appendixtocdepth}
\let\originalsection\section
\let\originalsubsection\subsection
\let\originalsubsubsection\subsubsection

\newif\ifinappendix
\inappendixtrue

\usepackage{glossaries}

\usepackage[utf8]{inputenc} 
\usepackage[T1]{fontenc}    
\usepackage{hyperref}       
\usepackage{cleveref}

\usepackage{url}            
\usepackage{booktabs}       
\usepackage{amsfonts}       
\usepackage{nicefrac}       
\usepackage{microtype}      

\input{math_commands.tex}

\input{sections_iclr/math}

\usepackage{hyperref}
\usepackage{url}
\usepackage{glossaries}
\usepackage{cleveref}
\input{sections_iclr/acronyms}

\title{Information Estimation with Discrete Diffusion}

\author{Alberto Foresti, Giulio Franzese \& Pietro Michiardi \\
Department of Data Science\\
EURECOM\\
450 Route des Chappes, 06410 Biot, France \\
\texttt{\{alberto.foresti,giulio.franzese,pietro.michiardi\}@eurecom.fr}
}

\iclrfinalcopy 
\begin{document}

\maketitle

\begin{abstract}
\input{sections_iclr/abstract}

\end{abstract}

\inappendixtrue

\section{Introduction}
\input{sections_iclr/introduction}

\section{Methodology}\label{sec:method}
\input{sections_iclr/klest}

\section{Our approach: \infosedd}\label{sec:infosed}
\input{sections_iclr/infosedd}

\section{Experiments}
\input{sections_iclr/experiments}
\subsection{Synthetic Experiments}\label{sec:synthetic}
\input{sections_iclr/synthetic}
\subsection{Application: Text Summarization}
\input{sections_iclr/text_experiments}
\subsection{Application: Genomics}
\input{sections_iclr/genomic_experiments}
\section{Conclusion}

\input{sections_iclr/conclusions}

\newpage

\bibliographystyle{plainnat}
\bibliography{sections_iclr/bibliography}

\newpage

\appendix
\inappendixtrue

\setcounter{section}{0}

\renewcommand{\section}[2][\relax]{%
  \ifx#1\relax
    \originalsection{#2}%
  \else
    \originalsection[#1]{#2}%
  \fi
  \ifinappendix
    \addcontentsline{apt}{section}{\protect\numberline{\thesection}#2}%
  \fi
}

\renewcommand{\subsection}[2][\relax]{%
  \ifx#1\relax
    \originalsubsection{#2}%
  \else
    \originalsubsection[#1]{#2}%
  \fi
  \ifinappendix
    \addcontentsline{apt}{subsection}{\protect\numberline{\thesubsection}#2}%
  \fi
}

\renewcommand{\subsubsection}[2][\relax]{%
  \ifx#1\relax
    \originalsubsubsection{#2}%
  \else
    \originalsubsubsection[#1]{#2}%
  \fi
  \ifinappendix
    \addcontentsline{apt}{subsubsection}{\protect\numberline{\thesubsubsection}#2}%
  \fi
}

\renewcommand{\cfttoctitlefont}{\Large\bfseries}
\renewcommand{\cftaftertoctitle}{\hfill}
\renewcommand{\cftbeforetoctitleskip}{20pt}
\renewcommand{\cftaftertoctitleskip}{20pt}

\clearpage
\begin{center}
\end{center}

\listofappendixtoc

\input{sections_iclr/appendix}

\end{document}

%% file: math_commands.tex
\usepackage{amsmath,amsfonts,bm}

\def\eqref#1{equation~\ref{#1}}

\def\1{\bm{1}}

\def\rvx{{\mathbf{x}}}
\def\rvy{{\mathbf{y}}}

\DeclareMathAlphabet{\mathsfit}{\encodingdefault}{\sfdefault}{m}{sl}
\SetMathAlphabet{\mathsfit}{bold}{\encodingdefault}{\sfdefault}{bx}{n}

\def\sN{{\mathbb{N}}}



%% file: sections_iclr/math.tex
\newcommand{\g}{\,|\,}

\newcommand{\prob}{\text{Pr}}

\DeclareRobustCommand{\KL}[2]{\ensuremath{\textsc{kl}\left[#1\;\|\;#2\right]}}

\newcommand{\bbR}{\mathbb{R}}

\newcommand{\defeq}{\stackrel{\text{\tiny def}}{=}}

\newcommand{\sub}[1]{{\texttt{\textit{\scriptsize {#1}}}}}

\newcommand{\pdata}{{p_\sub{\!d\!a\!t\!a}}}
\newcommand{\qdata}{{q_\sub{\!d\!a\!t\!a}}}

\newcommand{\pmarginal}{m}

\newcommand{\hamming}{d_{\text{hamming}}}

\newcommand{\fp}{\overset{\rightarrow}{p}}
\newcommand{\bp}{\overset{\leftarrow}{p}}

\newcommand{\fq}{\overset{\rightarrow}{q}}
\newcommand{\bq}{\overset{\leftarrow}{q}}

\newcommand{\fu}{\overset{\rightarrow}{u}}

\newcommand{\fratesp}{\overset{\rightarrow}{Q}}
\newcommand{\bratesp}{\overset{\leftarrow}{Q}}

\newcommand{\fratestokp}{{\overset{\rightarrow}{Q}}^{\text{tok}}}

\newcommand{\support}{\chi}

\newcommand{\fproc}{\overset{\rightarrow}{X}}
\newcommand{\bproc}{\overset{\leftarrow}{X}}

\newcommand{\fprocy}{\overset{\rightarrow}{Y}}

\newcommand{\xbar}{\bar{x}}

\newcommand{\pjump}{\prob(\fprocy_s\text{ jumps to }\absorb\text{ in }[0,t])}

\newcommand{\ones}{\mathbf{1}}

\newcommand{\expected}{\mathbb{E}}

\newcommand{\boperator}{\mathcal{B}}

\newcommand{\perr}{\epsilon_p}
\newcommand{\qerr}{\epsilon_q}

\newcommand{\scorep}{s^p_\theta}
\newcommand{\scoreq}{s^q_\phi}

\newcommand{\scorefn}{s_\theta}

\DeclareMathAlphabet{\pazocal}{OMS}{zplm}{m}{n}

\newcommand{\cumnoise}{\overline{\sigma}}
\newcommand{\absorb}{\emptyset}
\newcommand{\ratio}{\frac{1}{N(e^{\cumnoise(t)}-1)}}

\newcommand{\boltzmann}{k_b}
\newcommand{\pf}{\lambda}

\newcommand{\infosedd}{\textsc{info-sedd}}

\newcommand{\isedd}{\mathcal{E}}



%% file: sections_iclr/acronyms.tex
\newacronym{ODE}{\textsc{ode}}{Ordinary Differential Equation}
\newacronym{CTMC}{\textsc{ctmc}}{Continuous Time Markov Chain}
\newacronym{KL}{\textsc{kl}}{Kullback-Leibler}
\newacronym{MLP}{\textsc{MLP}}{Multi-Layer-Perceptron}
\newacronym{NLP}{\textsc{NLP}}{Natural Language Processing}
\newacronym{GWAS}{\textsc{GWAS}}{Genome-Wide Association Study}
\newacronym{WSSS}{\textsc{WSSS}}{Weakly Supervised Semantic Segmentation}
\newacronym{CAM}{\textsc{CAM}}{Class Activation Map}
\newacronym{SDE}{\textsc{SDE}}{Stochastic Differential Equation}
\newacronym{DTC}{\textsc{DTC}}{Dual Total Correlation}
\newacronym{TC}{\textsc{TC}}{Total Correlation}
\newacronym{MI}{\textsc{MI}}{Mutual Information}
\newacronym{DWDSE}{\textsc{dwdse}}{Diffusion Weighted Denoising Score Entropy}
\newacronym{GP}{\textsc{GP}}{Gaussian Process}
\newacronym{CI}{\textsc{CI}}{Confidence Interval}
\newacronym{MEC}{\textsc{MEC}}{Minimum Entropy Coupling}

%% file: sections_iclr/abstract.tex
Information-theoretic measures, such as \gls{MI}, play a crucial role in understanding non-linear relationships between random variables and are widely used across scientific disciplines. Yet, their use on real-world discrete data remains challenging. Existing methods typically rely on embedding discrete data into a continuous space and apply neural estimators originally designed for continuous distributions. This process requires careful engineering for both the embedding model and estimator architecture, but suffers from issues related to high data dimensionality. In this work, we introduce \infosedd, a discrete diffusion–based approach that bridges information-theoretic estimation and generative modeling such that they can be used to compute Kullback-Leibler divergences. Backed by Continuous Time Markov Chains theory principles, the design of \infosedd\ is lightweight and scalable and allows seamless integration with pretrained models. 
We showcase the versatility of our approach through applications on motif discovery in genetic promoter data, semantic-aware model selection in text summarization, and entropy estimation in Ising models.
Finally, we construct consistency tests on real-world textual and genomics data. Our experiments demonstrate that \infosedd\ outperforms alternatives that rely on the ``embedding trick''. Our results position \infosedd\ as a robust and scalable tool for information-theoretic analysis of discrete data. We provide the open source code at \url{https://github.com/AlbertoForesti/mutinfo-diffusion}.

%% file: sections_iclr/introduction.tex
Information theoretic measures represent a powerful tool to understand non-linear relationships between random variables \citep{shannon1948mathematical, mackay2003information} and find a wide range of applications in scientific fields \citep{Karbowski_2024, Eckford_2016}. Mutual information (MI), in particular, has become an established metric in machine learning \citep{bell1995information, stratos2018mutual, belghazi2018mine, oord2019representation, hjelm2019learning}, both for training models \citep{alemi2019deep, chen2016infogan, zhao2018information} and at inference time \citep{alemi2019gilbo, huang2020evaluating}.

Estimating information theoretic quantities remains an open problem, and different paradigms for their estimation have emerged. Classical parametric and non-parametric methods \citep{pizer1987adaptive, moon1995estimation, kraskov2004estimating, gao2015efficient} have been recently superseded by variational approaches \citep{barber2004algorithm, nguyen2007neurips, nowozin2016neurips, poole2019variational, wunder2021reverse, letizia2024mutual, federici2023effectiveness} and neural estimators. Notably, several approaches rely on optimizing a lower bound \citep{papamakarios2017masked, belghazi2018mine, oord2019representation, song2019, rhodes2020telescoping, letizia2022copula, brekelmans2023improving}. However, \citet{mcallester2020formal,song2019understanding} challenge the applicability of this kind of estimator, especially in high \gls{MI} scenarios, as they require an amount of samples that grows exponentially with the ground truth \gls{MI}. To overcome this limitation, a new class of estimators has emerged \citep{franzese2023minde,butakov2024mutual,kholkin2025infobridgemutualinformationestimation}. Unfortunately, these methods focus on continuous data and, despite their practical importance, few estimators for high-dimensional \textbf{discrete} distributions have been proposed in the literature. While classical estimators for discrete random variables exist \citep{pinchas2024comparative}, their accuracy rapidly decreases with increasing data dimensionality. Applications that would benefit from scalable estimators of \gls{MI} include, among others, DNA or peptide sequencing \citep{newcomb2021use, xiaadanovo}, text summarization \citep{darrin2024textttcosmicmutualinformationtaskagnostic} and neuroscience \citep{chai2009exploring}, to name a few examples. Consequently, the development of new estimation techniques is of paramount importance for the broader scientific community.

A common workaround to deal with high-dimensional, discrete data is to embed it in a continuous space and use neural estimators conceived for continuous distributions. One recent example is \citep{lee2024benchmarksuiteevaluatingneural}, where it is shown that the embeddings of pretrained language models can provide meaningful representations to estimate information theoretic quantities in unstructured data. However, such a process may not fully capture the discrete nature of the underlying data and might suffer from several limitations, such as the necessity to consider application-specific embeddings. For example, we show that \textsc{minde} \citep{franzese2022}, which is a strong \gls{MI} estimator in continuous domains, like image data, struggles with discrete data. 

In this work, we build on the growing literature of diffusion-based estimators \citep{kong2023information, franzese2023minde}, and present \infosedd, a novel method for estimating information theoretic quantities of discrete data using \glspl{CTMC} \citep{lou2024discrete}. These stochastic processes have recently seen a surge in popularity for applications such as generative language modeling \citep{lou2024discrete, sahoo2025simple, nie2025large}. Their fundamental working principle is the reversal of a perturbation process which starts with clean data from a given distribution and that converges to uninformative noise. The workhorse of these approaches is the \textit{score function}, which contains information about the probability distributions associated with the \glspl{CTMC} at different time instants. Our proposed method, \infosedd, builds upon such mathematical framework, extending it via Dynkin's lemma \citep{hanson2007applied}, and leverages score functions to compute key information-theoretic metrics, such as \gls{MI} between two random variables, and the entropy of a given distribution. By carefully selecting perturbation processes, our approach requires training only a single parametric model to compute \gls{MI} across arbitrary subsets of variables. Furthermore, \infosedd\ seamlessly integrates with pretrained models, without requiring ad-hoc procedures to work with discrete data.

To rigorously evaluate our method, we perform a large number of experiments both on synthetic and real data, and compare \infosedd\ to state-of-the-art methods. We design a synthetic benchmark with ground truth \gls{MI} that presents challenges such as high data dimensionality and high \gls{MI} scenarios. Our results demonstrate that \infosedd\ is both robust and consistently outperforms existing estimation methods. We then focus on two application domains in which we compare \infosedd\ to competitors, which all require the ``embedding trick'' discussed above. First, we tackle the text summarization domain, evaluate the consistency of \gls{MI} estimators, and study if \gls{MI} represents a meaningful signal to perform model selection, as \infosedd\ estimates are well aligned with human metrics for evaluating text summarization. We also present results in the domain of genomics, whereby we evaluate the consistency of \gls{MI} estimators, and show that \infosedd\ can be used to detect motif location in DNA sequencing data.
Overall, our results indicate that \infosedd\ outperforms alternatives for \gls{MI} estimation, and pave the way to exploit information measures as meaningful signals for a variety of downstream tasks involving high-dimensional discrete data.

%% file: sections_iclr/klest.tex
We begin by defining the relation between \glspl{CTMC} over discrete state spaces \citep{anderson2012continuous} and the computation of \gls{KL} divergences. First, in \Cref{subsec:ctmc}, we provide a brief introduction to the fundamentals of \glspl{CTMC}, emphasizing their time-reversal properties and parametric approximations \citep{lou2024discrete}. Then, in \Cref{subsec:klfor}, we demonstrate how these processes can be adapted for divergence estimation, specifically by analyzing two processes that share the same generator but differ in their initial conditions.

\subsection{Preliminaries}\label{subsec:ctmc}
Consider a \gls{CTMC} $\fproc_t, t \in[0,T]$, defined over a finite state space $\support = \left\{1,\hdots,N\right\}$ and  specified by the infinitesimal generators $\fratesp_t: [0,T]\rightarrow \mathbb{R}^{N\times N}$, where the diagonal entries satisfy $\fratesp_t(a,a)=-\sum_{a\neq b}\fratesp_t(a,b)$, with $\fratesp_t(a,b)\geq 0, a\neq b$. As established by \citet{anderson2012continuous}, the time evolution of the probability distribution $Pr(X_t=x)\defeq \fp_t(x)$ satisfies the following \gls{ODE}: 
\(\fp_t =\fp_0+\int_0^t \fratesp_s\fp_s ds\),
where the initial conditions of the process $\fp_0$ determine the distribution $\fp_t$ at any time $t$. A key property of \glspl{CTMC} is that their time-reversed counterpart, recently used for generative modeling purposes \citep{lou2024discrete}, also follows a \gls{CTMC}, but with a different set of transition matrices. More precisely, defining the time-reversed process as $\bp_t\defeq\fp_{T-t}$, the reverse process evolves according to the following \gls{ODE} \citep{lou2024discrete,sun2023scorebasedcontinuoustimediscretediffusion}:
\(\bp_t =\fp_T+\int_0^t \bratesp_s\bp_s ds,\)
where the reverse-time transition matrices $\bratesp_t$ relate to the forward transition matrices as follows:

{\small \begin{align}\label{eq:ratesdef}
    \bratesp_t(b,a) &= \left(\tfrac{\fp_{T-t}(b)}{\fp_{T-t}(a)} \fratesp_{T-t}(a,b)\right) (1-\delta(a,b))+\left(-\sum_{b\neq a}\bratesp_t(b,a)\right)\delta(a,a).
\end{align}}

Under appropriate technical conditions on $\fratesp_t$ \citep{lou2024discrete}, the terminal distribution $\fp_T$ converges to a known reference distribution $\pi$, which is independent of the initial distribution $\fp_0$. This property enables sampling from $\fp_0$ by simulating a \gls{CTMC} with appropriately chosen generators \citep{sun2023scorebasedcontinuoustimediscretediffusion, kelly1981reversibility}. However, other than simple and uninteresting scenarios, exact knowledge of the quantities $\tfrac{\fp_{T-t}(b)}{\fp_{T-t}(a)}$ is out of reach. A practical solution is to substitute in this numerical integration a parametric function $\scorep(a,t)_b$, whose parameters are optimized according to the \gls{DWDSE} loss function (Equation 10, Section 3.2 of \citet{lou2024discrete}). 
Whenever the context is clear, we simplify the notation for the parametric score in the remainder of the paper and denote it with $\scorep(\fproc_t)_x$ instead of $\scorep(\fproc_t,t)_x$.

\subsection{\gls{KL} Divergences via \glspl{CTMC}}\label{subsec:klfor}

Next, we show how to extend the \gls{CTMC} framework to compute the \gls{KL} divergence between two probability distributions $\fp_0$ and $\fq_0$ defined over the same support $\support$, expressed as $\KL{\fp_0}{\fq_0}$. 
To achieve this, we construct two Markov chains that differ only in their initial conditions: one initialized from $\fp_0$, and the other initialized from $\fq_0$. The \gls{KL} divergence can be expressed as: 

\begin{equation}\label{eq:kl_expressions}
    \KL{\fp_0}{\fq_0} = \expected\left[\log\tfrac{\fp_0}{\fq_0}(\bproc_T)\right] =\expected\left[\log\tfrac{\bp_T}{\bq_T}(\bproc_T)\right] = \expected \left[\expected\left[\log\tfrac{\bp_T}{\bq_T}(\bproc_T)\bigg|\bproc_0\right]\right].
\end{equation}

The last term in \Cref{eq:kl_expressions} can be rewritten using Dynkin's formula \citep{hanson2007applied}, which states that for a generic function \( f:\support\times[0,T]\to\bbR \), we have:

\begin{equation}\label{eq:dynkin}
    \expected\left[f(\bproc_T,T)\bigg|\bproc_0\right]- f(\bproc_0,0)=\expected\left[\int_0^T\tfrac{\partial f}{\partial t}(\bproc_t,t)+\boperator[f](\bproc_t,t)dt\bigg|\bproc_0\right],
\end{equation}

where \( \boperator \) is the \textit{backward operator}, which we define as $  \boperator[f](a,t) = \sum_{b\neq a}\bratesp_t(b,a)(f(b)-f(a))$.


By combining the result from \Cref{eq:dynkin} with \Cref{eq:kl_expressions}, the \gls{KL} divergence between discrete distributions can be conveniently approximated as:
\begin{small}
\begin{equation}\label{eq:infosedd}
\KL{\fp_0}{\fq_0}\approx\expected\left[\int_0^T\sum_{x\neq \fproc_t} \fratesp_{t}(\fproc_t,x)\left(K\left(\tfrac{\fp_t(x)}{\fp_t(\fproc_t)}\right)+\tfrac{\fq_t(x)}{\fq_t(\fproc_t)}-\tfrac{\fp_t(x)}{\fp_t(\fproc_t)}\log\tfrac{\fq_t(x)}{\fq_t(\fproc_t)}\right)dt\right],
\end{equation}
\end{small}
where \( K(a)=a(\log(a)-1) \). We omit the term \( \expected\left[\log{\tfrac{\bp_0}{\bq_0}}(\bproc_0)\right] \), as both \( \bp_0 \) and \( \bq_0 \) converge to \( \pi \) \citep{lou2024discrete}. 

The key quantities \( \tfrac{\fp_{t}(b)}{\fp_{t}(a)} \) and \( \tfrac{\fq_{t}(b)}{\fq_{t}(a)} \) in the expression above are not directly accessible. To address this, we substitute these ratios with parametric approximations optimized via \gls{DWDSE} loss, leading to the following \gls{KL} estimator: 
\begin{small}
\begin{equation}\label{eq:infosedd-param} 
\KL{\fp_0}{\fq_0} \approx \expected\left[\int_0^T\sum_{x\neq \fproc_{t}} \fratesp_{t}(\fproc_{t},x)\left(K\left(\scorep(\fproc_{t})_x\right)+\scoreq(\fproc_{t})_x-\scorep(\fproc_{t})_x\log\scoreq(\fproc_{t})_x\right) dt\right].
\end{equation}
\end{small}

Estimating \Cref{eq:infosedd-param} using Monte Carlo techniques is straightforward: we sample time instants \( t \) uniformly in \( [0,T] \), simulate the forward process \( \fproc_t \), and compute the required quantities using the parametric scores. 
\Cref{eq:infosedd-param} enables estimating information measures, such as \gls{MI} and Entropy. 
However, the \gls{KL} estimator in its general form is not scalable for practical purposes: our practical method, described next, addresses this limitation.

%% file: sections_iclr/infosedd.tex
We are now ready to introduce our \gls{MI} estimator, \infosedd, which is based on \Cref{eq:infosedd-param}. Given two random variables, $X$ and $Y$, we propose two approaches: 

\begin{enumerate}
    \item \textbf{Joint Method} [\infosedd-\textsc{j}]: \gls{MI} can be expressed in terms of the \gls{KL}-divergence between the joint distribution $p_{XY}$ and the product of the marginals $\pmarginal_{XY} = p_X \otimes p_Y$, $I(X,Y) = \KL{p_{XY}}{\pmarginal_{XY}}$, where $\otimes$ denotes the Kronecker product.
    \item \textbf{Conditional Method} [\infosedd-\textsc{c}]: alternatively, \gls{MI} can be expressed as the \gls{KL} divergence between the conditional distribution and the marginal distribution, $I(X,Y) = \expected[\KL{p_{Y|X}}{p_{Y}}]$.
\end{enumerate}

In high dimensional applications, a naive implementation of \Cref{eq:infosedd-param} quickly becomes unfeasible. Indeed, the size of the number of entries of the matrix $\fratesp_t$ scales with $|\support|^2$, which is intractable. 
However, in many cases of interest, the random variables can be naturally decomposed into a structured sequence of $D$ subcomponents, each taking values from a discrete set of size $|\chi|$ \citep{lou2024discrete, austin2021structured, campbell2022continuous}, i.e. $X=[X_1,\dots,X_D]$, leading to a total state space of size $|\support|^D$. This structured decomposition enables the use of sparse rate matrices, which constrain the \gls{CTMC} to modify only one subcomponent at a time, significantly reducing computational complexity. This allows us to consider sequences which only differ in one component when computing the entries of $\fratesp$, or, equivalently, sequences with unit Hamming distance.\footnote{This prohibits any sequence to jump to another sequence with Hamming distance larger than one, but only in the infinitesimal time regime $dt$. Once we consider longer time intervals, perturbations immediately become non-local. Empirical evidence in the field of discrete diffusion models indicates that such decomposition allows dealing with complex distributions with non-local interactions, such as text/language and single-cell RNA sequencing data \citep{sahoo2025simple, lou2024discrete}} In particular, the non-zero entries of $\fratesp$ are determined by a shared $|\chi|\times|\chi|$ local rate matrix $\fratestokp$. Specifically, if two sequences $a$ and $b$ differ only at the $i$-th component, then their transition rate is given by \(\fratesp_t(a,b)=\fratestokp_t(a_i,b_i)\). 



Expressing the \gls{MI} as the \gls{KL} divergence between the joint distribution and the product of marginals, $\KL{p_{XY}}{\pmarginal_{XY}}$, requires training two separate score models, each tailored to a specific distribution. A carefully chosen transition matrix can circumvent this requirement, allowing for a single model to be trained instead. In particular, selecting $\fratestokp_t=\sigma(t)\fratestokp_{\text{absorb}}$, with $\sigma(t)$ a fixed scalar function, and the absorbing matrix $\fratestokp_{\text{absorb}}$    
as defined in \citet{lou2024discrete, campbell2022continuous, austin2021structured}, 
ensures that the subcomponents can only transition into an absorbing state $\absorb$. 
This choice is crucial because it enables the computation of marginal scores using a model trained solely on the joint distribution: 
\begin{equation}\label{eq:free_marginal}
    \tfrac{\fp_t(\fproc_t=x,\fprocy_t=\absorb)}{\fp_t(\fproc_t=x',\fprocy_t=\absorb)}=\tfrac{\fp^X_t(x)}{\fp^X_t(x')},\quad \tfrac{\fp_t(\fproc_t=\absorb,\fprocy_t=y)}{\fp_t(\fproc_t=\absorb,\fprocy_t=y')}=\tfrac{\fp^Y_t(y)}{\fp^Y_t(y')}.
\end{equation}

For the mathematical proof the reader is invited to check \Cref{subsec:free_marginal}. This result implies that a single score model trained on the joint distribution is sufficient for computing the marginal scores as well.\footnote{This configuration adds an absorbing state, increasing the dimension of the support} By integrating these design choices, we present the pseudocode for \infosedd-\textsc{j} and \infosedd-\textsc{c} in \Cref{app:pseudocode}, in  \Cref{alg:infosedd-j} and \Cref{alg:infosedd-c} respectively.

\noindent \textbf{Theoretical properties.} We investigate the theoretical properties of \infosedd, specifically focusing on the error decomposition and consistency. We establish that under mild assumptions on the boundedness of the score functions (constants $C_1, C_2$) and the neural network approximation errors ($\perr, \qerr$), the deviation of the \infosedd\ estimator from the true \gls{KL} divergence is bounded by:

\begin{equation}\label{eq:main_bound_body}
    \left| \expected_{x\sim p}\isedd(\scorep,\scoreq;x) - \KL{p}{q} \right| 
    \leq 
    \underbrace{\bar{\sigma}(T)D|\chi|\left(1+\tfrac{C_2}{C_1}\right)(\perr+\qerr)}_{\text{Estimation Error}} 
    + 
    \underbrace{(1-\fp_T(\absorb^D)) (D C_2 \log |\chi|)}_{\text{Truncation Bias}}.
\end{equation}

Where \(\isedd(\scorep,\scoreq;x)\) corresponds to the right-hand side of \Cref{eq:infosedd-param}.
This bound highlights a key trade-off: the first term, representing the estimation error, scales linearly with the score error and constants dependent on the data distribution. The second term represents a truncation bias arising from the finite time horizon $T$, which vanishes exponentially as the probability of the absorbing state $\fp_T(\absorb^D)$ approaches 1. Consequently, \infosedd\ is a consistent estimator up to this exponentially decaying bias, allowing for accurate \gls{KL} estimation without the exponential variance associated with standard importance sampling methods. We refer the reader to \Cref{app:theory} for the full derivation and proofs.

\noindent \textbf{Estimating entropy.} Our method can also be used for Entropy estimation, since it can be expressed in terms of the \gls{KL} divergence between a distribution $\fp_0$ and the uniform distribution $\fu_0$, as follows: $H(\fp_0) = \log N -\KL{\fp_0}{\fu_0}$. Since the ratio $\frac{\fu_t(x)}{\fu_t(\absorb)}=\ratio$, with $\cumnoise(t)=\int_0^t\sigma(s)ds$ (see \cref{subsec:absorb_ratio}), we can adapt the formulation in \Cref{eq:infosedd-param} to derive \textsc{info-sedd-h}, an Entropy estimator, which is detailed in \Cref{app:pseudocode}, in \Cref{alg:compute_entropy}.

%% file: sections_iclr/experiments.tex
Next, we evaluate \infosedd\ on a series of synthetic experiments, as well as on several realistic applications. Overall, while our competitors struggle with high \gls{MI} scenarios and high-dimensional data, \infosedd\ obtains accurate estimates. Using text and DNA data, we show that \infosedd\ is not only accurate, but also useful for various downstream tasks. In \Cref{app:additional_results}, we provide additional results, showing that \infosedd\ can produce accurate Entropy estimates in Ising models \citep{onsager1944crystal}, which are used to understand the state of a particle lattice using their spins.

%% file: sections_iclr/synthetic.tex
We validate our method on synthetic distributions with known mutual information (full details are in \Cref{app:synthetic}). 
Given two vectors $X=(x_1, x_2, \ldots,x_{D})$ and $Y=(y_1, y_2, \ldots,y_{D})$ sampled from their respective discrete distributions, we denote vector dimensionality by $D$, and the support $|\chi|$ as the number of discrete values each element $x_i$, $y_i$ can take. 

Next, we benchmark \infosedd\ and several competitors on a high dimensional synthetic benchmark, by increasing both \gls{MI} and $D$ as shown in \Cref{tab:mi_comparison_std}. We compare the results of our proposed methodology against some of the estimators from \citet{letizia2024mutual}, who introduce data derangements to improve bias and variances of existing variational estimators, including \textsc{nwj} \citep{nguyen2007neurips}, \textsc{smile} \citep{song2019understanding} and \textsc{mine} \citep{belghazi2018mine}. \citet{letizia2024mutual} also proposes the \textsc{f-dime} estimators \textsc{gan-dime}, \textsc{kl-dime}, \textsc{hd-dime}, which are shown to work better at high \gls{MI}. We also benchmark \textsc{minde} \citep{franzese2023minde}, a generative neural estimator for continuous data. We use the same backbone for all methods (see \Cref{app:synthetic}), with only minor tweaks to initial and final layers to accommodate the specifics of each method. For our method, we report the results with the \infosedd\textsc{-j} variant, trained using the \gls{DWDSE} loss. We train all methods using $10^5$ samples and a batch size of $1024$ for $10^5$ steps.

\input{tables/synthetic/high_dim_error_bars}

Results in \Cref{tab:mi_comparison_std} demonstrate that \infosedd\ consistently outperforms competing methods. \infosedd\ performs better with high dimensional data and high \gls{MI}, while existing neural estimators  fail. This is not the only scenario where \infosedd\ shines, as demonstrated extensively in \Cref{app:synthetic}. In \Cref{app:big_supp} we perform an ablation study on $|\chi|$, showing that \infosedd\ is robust to changes in support size, where, instead, the competitors cannot provide accurate estimates. Accuracy, however, is not the only dimension where competitors struggle. In \Cref{app:convergence}, we show that 
even when competing estimators like \textsc{gan-dime} and \textsc{smile} provide correct estimates, they take more epochs to converge compared to \infosedd, resulting in slower training. Finally, in \Cref{app:sample_complexity} we evaluate empirically the sample complexity of \infosedd\ 
observing that it is accurate even when using only $10^3$ samples.

Overall, our complete set of synthetic experiments confirms that \infosedd\ is robust, efficient, and parsimonious in terms of sample size. Next, we revert to realistic use cases to further evaluate information-theoretic estimators in even more challenging scenarios.

%% file: tables/synthetic/high_dim_error_bars.tex
\begin{center}
\captionof{table}{Results for the high dimensional synthetic benchmark for all estimators with given \gls{MI} and same vector length (D) for each modality. We report the mean estimate and standard deviation over 10 seeds. Best estimates are marked in bold.}
\tiny
\begin{tabular}{lllllllll}
\toprule
Estimator & \infosedd & \textsc{gan-dime} & \textsc{hd-dime} & \textsc{kl-dime} & \textsc{minde} & \textsc{mine} & \textsc{nwj} & \textsc{smile} \\
 &  &  &  &  &  &  &  &  \\
\midrule
MI=10, D=10 & \textbf{9.92} $\pm$ 0.12 & 12.15 $\pm$ 0.89 & 9.73 $\pm$ 0.43 & 8.38 $\pm$ 0.90 & 14.01 $\pm$ 2.91 & 10.21 $\pm$ 6.33 & 6.16 $\pm$ 2.11 & 12.83 $\pm$ 0.95 \\
MI=20, D=20 & \textbf{20.02} $\pm$ 0.21 & 22.09 $\pm$ 1.75 & 12.65 $\pm$ 1.07 & 7.51 $\pm$ 0.56 & 26.98 $\pm$ 3.16 & 8.82 $\pm$ 0.80 & 6.50 $\pm$ 0.84 & 23.11 $\pm$ 1.41 \\
MI=30, D=30 & \textbf{29.83} $\pm$ 0.54 & 20.74 $\pm$ 1.75 & 11.72 $\pm$ 2.69 & 7.02 $\pm$ 0.43 & 31.08 $\pm$ 4.33 & 7.41 $\pm$ 1.23 & 6.35 $\pm$ 0.34 & 21.79 $\pm$ 1.08 \\
MI=40, D=40 & \textbf{39.11} $\pm$ 0.65 & 19.64 $\pm$ 1.33 & 11.68 $\pm$ 0.94 & 6.52 $\pm$ 0.32 & 33.97 $\pm$ 3.32 & 6.91 $\pm$ 0.66 & 6.24 $\pm$ 0.59 & 20.13 $\pm$ 1.27 \\
MI=50, D=50 & \textbf{47.77} $\pm$ 1.18 & 17.27 $\pm$ 1.46 & 10.47 $\pm$ 1.12 & 6.41 $\pm$ 0.62 & 32.60 $\pm$ 3.93 & 7.21 $\pm$ 1.14 & 5.95 $\pm$ 0.31 & 18.97 $\pm$ 1.05 \\
\bottomrule
\end{tabular}
\label{tab:mi_comparison_std}
\end{center}

%% file: sections_iclr/text_experiments.tex
We consider the task of text summarization, which aims at generating concise document summaries, either by directly extracting sentences from the original corpus, or by introducing new, relevant phrases not present in the original text \citep{nallapati2016abstractive}. We validate \infosedd\ using the \textsc{SummEval} dataset \citep{fabbri2021summeval} which assembles collections of summaries generated by 23 different models, providing human judgments for outputs generated by 15 of these models. We generate a collection of datasets of texts/summaries pairs from the \textsc{SummEval} dataset, such that we can scramble part of such pairs to gauge \gls{MI} between texts and summaries distributions. 
We compare \infosedd\ and alternative methods, showing how much each estimator is \textit{consistent} with a ``theoretical'' trend of \gls{MI}. 
We also study the application of \infosedd\ for summarization model selection, by estimating the \gls{MI} of each collection of model outputs with respect to the reference texts. Studying how these estimates compare to the human evaluations of the summaries helps practitioners to identify which human metric is best aligned to \gls{MI}. For more details, refer to \Cref{app:summarization}. 
\paragraph{Consistency test.} We perform a consistency test for \gls{MI} estimation between model-generated summaries and reference texts: we select the summaries generated by \textsc{bart} \citep{lewis2020bart} and pair them with the original text with a probability $\rho$ and with a random text in the dataset with probability $1-\rho$. While we cannot establish an exact ground truth for these experiments, we can determine that \gls{MI} should grow linearly as a function of $\rho$, under the assumption of \gls{MI} being significantly larger than $\log2$. 
This assumption is backed by studies \citep{takahira2016entropy, cover1978convergent} that estimate the entropy rate of textual data to be larger than 1 bit per character (bpc), and by works that estimate \gls{MI} in a similar context \citep{darrin2024textttcosmicmutualinformationtaskagnostic}. To support our claims and provide an order-of-magnitude estimate,  we multiply the entropy rates of \citet{takahira2016entropy} and \citet{cover1978convergent} with the average length of dataset summaries, to obtain entropy estimates for English texts with the same length, obtaining values of $256$ nats and $303$ nats, respectively. For a more detailed discussion, refer to \Cref{app:summarization}. 
We use the \textsc{mdlm-small} model \citep{sahoo2025simple} as the backbone, with minimal changes to the architecture to accommodate our competitors. We also slightly modify the training strategy of \citet{sahoo2025simple} to allow a more efficient learning of marginal and conditional scores for \infosedd\ (check \Cref{app:summarization} for details). This allows us to keep the same context length, architecture, and a similar number of parameters for all methods. Note that for all competitors, we project text tokens into an embedding space of fixed dimension, by jointly learning an embedding look-up table for each method. For more experimental details, check \Cref{app:summarization}.

        

\begin{figure}[htbp]
    \centering
    \begin{minipage}[b]{0.48\textwidth} 
        \centering
        \includegraphics[width=\textwidth]{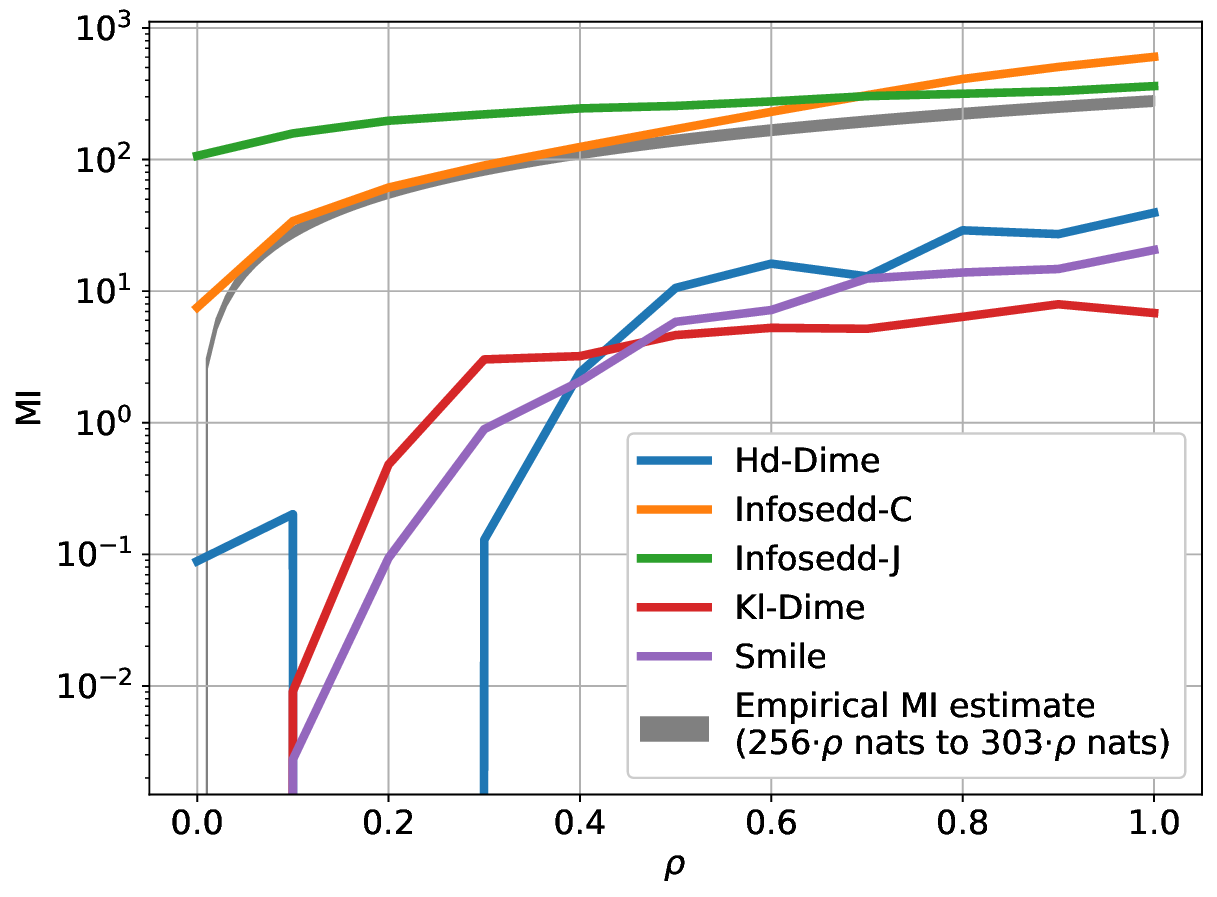}
        \captionsetup{justification=centering}
        \caption{Consistency results for selected estimators.}
        \label{fig:consistency-summaries}
    \end{minipage}\hfill
    \begin{minipage}[b]{0.48\textwidth} 
        \centering
        \includegraphics[width=\textwidth]{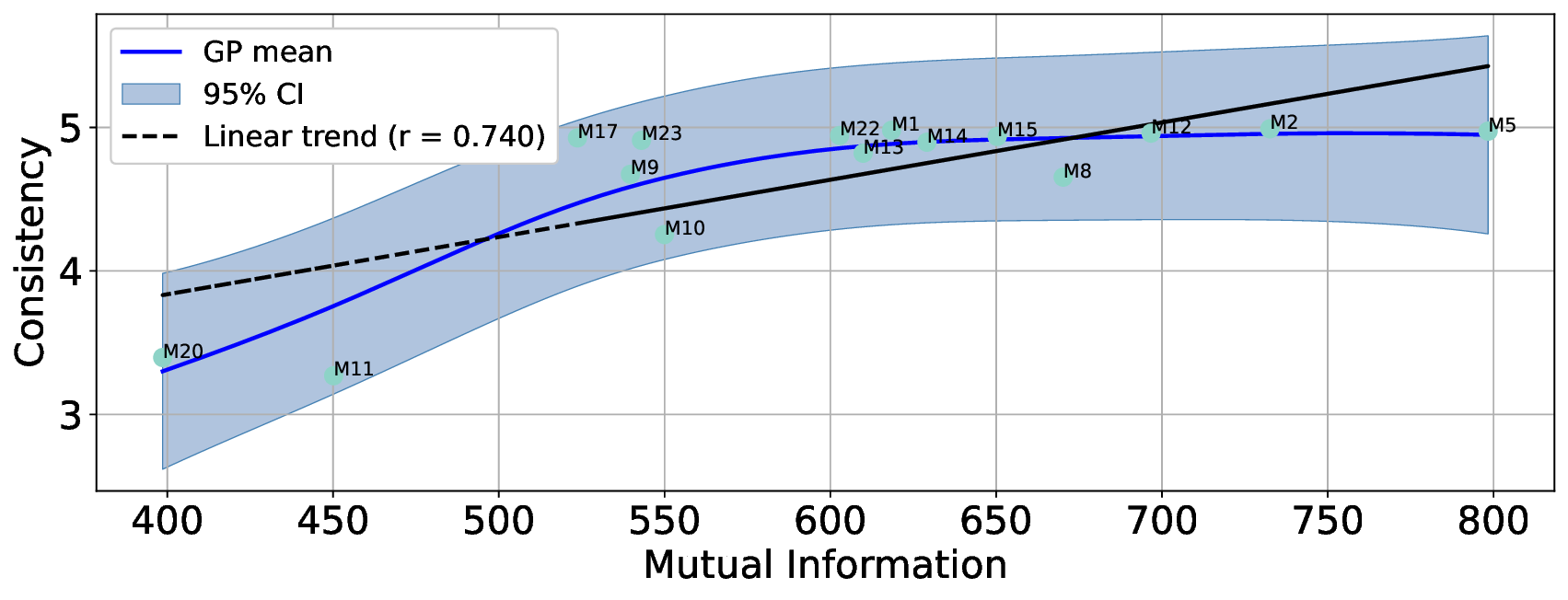}
        \captionsetup{justification=centering}
        \caption{Consistency human metric compared to \gls{MI} as estimated by \infosedd-\textsc{c}.}
        \label{fig:hm-consistency-infoseddc}
        
        \includegraphics[width=\textwidth]{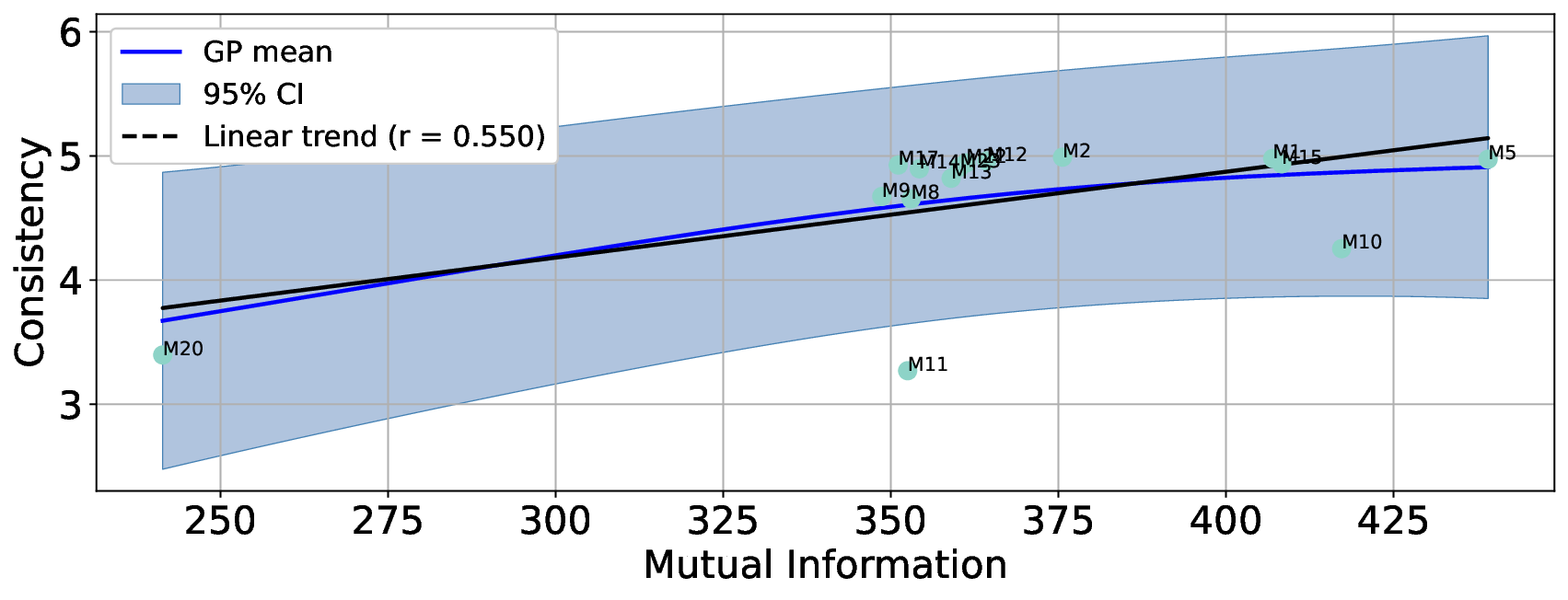}
        \captionsetup{justification=centering}
        \caption{Consistency human metric compared to \gls{MI} as estimated by \infosedd-\textsc{j}.}
        \label{fig:hm-consistency-infoseddj}
    \end{minipage}
\end{figure}

As shown in \Cref{fig:consistency-summaries}, where we report \gls{MI} estimates for $\rho=0.0,0.1,...,1.0$, the consistency of both \infosedd\ variants is the highest among all competitors, closely matching the empirical derivation outlined above. Note that \infosedd-\textsc{c} obtains \gls{MI} estimates closer to zero than the joint variant, when $\rho=0.0$. 
Instead, variational approaches cannot reliably estimate \gls{MI} with values larger than the logarithm of the batch size used for estimating it \citep{mcallester2020formal, song2019understanding}. This results in \gls{MI} being underestimated, a problem that could only be addressed by impractical batch sizes. 
Despite \textsc{kl-dime} and \textsc{smile} exhibit an approximately linear correlation, they obtain low \gls{MI} values, incompatible with the empirical derivation. \textsc{hd-dime} can achieve higher values of \gls{MI}, but fails for $\rho<0.5$. \textsc{minde} does not provide meaningful \gls{MI} estimates: this is likely due to the high embedding dimensionality which, together with the sequence length, is a challenging scenario for continuous score models.


\paragraph{Model selection.} We now evaluate if \gls{MI} is a meaningful signal to perform model selection, by measuring if \gls{MI} estimates are compatible with human-preference metrics, as done by \citet{darrin2024textttcosmicmutualinformationtaskagnostic}.
This is done by estimating the \gls{MI} of the text-model summary datasets in \textsc{SummEval} \citep{fabbri2021summeval}, for every model with available human metrics. The human metrics we consider are: \textbf{coherence}, which measures how well-structured and coherent the summary is; \textbf{consistency}, which quantifies how much the statements in the summary are entailed by the source; \textbf{fluency}, which evaluates the grammatical correctness and the formatting of the summary; and \textbf{relevance}, which measures if the summary captures only the most important information of the source text.
\Cref{fig:corr-hm-mi}, which shows Pearson correlation and Kendall's Tau test results, indicates that \gls{MI} correlates the most with consistency. This is expected, as consistency effectively quantifies how much information the text and summary share. Note, however, that raters were asked to penalize hallucinated facts present in the summaries, which is not captured by \gls{MI}. The second metric which correlates the most with \gls{MI} is fluency. The definition of fluency does not take into account a comparison between summary and source text, but it also strongly correlates with consistency. Thirdly, \gls{MI} correlates well with relevance. Relevance considers a comparison between summary and source text in terms of what is considered important information by human annotators. However, important information is highly subjective, and human intentions are likely not captured by \gls{MI}. Lastly, coherence does not quantify a relationship between source text and summaries, which explains the low correlation with \gls{MI}. 
In \Cref{fig:hm-consistency-infoseddc,fig:hm-consistency-infoseddj} we show the relation between the \gls{MI} estimates of \infosedd-\textsc{c} and \infosedd-\textsc{j} and consistency by running a \gls{GP} regression with a Matérn Kernel with smoothing parameter $\nu=1.5$ and noise level $0.01$. We report the \gls{GP} mean, its 95\% confidence interval and the Pearson correlation. \Cref{fig:hm-consistency-infoseddc,fig:hm-consistency-infoseddj} show that consistency saturates around the maximum score, whereas \gls{MI} does not have the same ceiling effect. The reader is referred to \Cref{app:summarization} for a similar discussion about the other metrics.
Overall, our results are in line with the findings by \citet{darrin2024textttcosmicmutualinformationtaskagnostic}, where \infosedd-\textsc{c} achieves a comparable correlation with the most important metric, consistency, albeit without requiring elaborate embedding models, that rely on a large set of parameters to be learned.

\begin{figure}[htbp]
    \centering
    {\scriptsize 
    \captionof{table}{Pearson (left) and Kendall's Tau (right) correlation of human metrics and \gls{MI} for \infosedd\ and selected competitors. The labels are shorthand notations for the human metrics: COH=coherence, CON=consistency, FLU=fluency, REL=relevance, OVR=overall. The reader is invited to check \Cref{app:corr_mats} for correlations with other metrics offered by the \textsc{SummEval} framework.}
    \label{fig:corr-hm-mi} 
    \begin{minipage}{0.48\linewidth}
        \centering
        \input{tables/pearson_correlations_all}
    \end{minipage}}
    \hfill
    {\scriptsize\begin{minipage}{0.48\linewidth}
        \centering
        \input{tables/kendall_correlations_all}
    \end{minipage}}
    
\end{figure}


%% file: tables/pearson_correlations_all.tex
\begin{tabular}{lrrrrr}
\toprule
 & \textsc{coh} & \textsc{con} & \textsc{flu} & \textsc{rel} & \textsc{ovr} \\
\midrule
\infosedd\textsc{-c} & 0.209 & 0.740 & 0.679 & 0.411 & 0.568 \\
\infosedd\textsc{-j} & -0.091 & 0.550 & 0.455 & 0.288 & 0.322 \\
\textsc{kl-dime} & 0.170 & 0.214 & 0.194 & 0.076 & 0.193 \\
\textsc{hd-dime} & -0.243 & 0.331 & 0.281 & -0.145 & 0.063 \\
\textsc{smile} & -0.367 & -0.074 & -0.162 & -0.149 & -0.221 \\
\bottomrule
\end{tabular}

%% file: tables/kendall_correlations_all.tex
\begin{tabular}{lrrrrr}
\toprule
 & \textsc{coh} & \textsc{con} & \textsc{flu} & \textsc{rel} & \textsc{ovr} \\
\midrule
\infosedd\textsc{-c} & 0.105 & 0.505 & 0.134 & 0.200 & 0.219 \\
\infosedd\textsc{-j} & 0.048 & 0.486 & 0.153 & 0.219 & 0.238 \\
\textsc{kl-dime} & 0.105 & 0.429 & 0.096 & 0.048 & 0.067 \\
\textsc{hd-dime} & -0.162 & -0.029 & 0.057 & -0.067 & -0.048 \\
\textsc{smile} & -0.238 & -0.105 & 0.019 & -0.067 & -0.086 \\
\bottomrule
\end{tabular}

%% file: sections_iclr/genomic_experiments.tex
The genomics field heavily relies on computational methods to improve our understanding of hidden patterns in large and complex genomic data sets from basic and clinical research projects \citep{libbrecht2015machine, whalen2022navigating, teschendorff2025epigenetic}.
We consider two fundamental problems in genetics (genome classification and motif identification in genetic promoters) that rely on DNA sequencing data: DNA is a nucleotide made of four types of nitrogen bases, Adenine (A), Thymine (T), Guanine (G), and Cytosine (C). The order, or sequence, of these bases determines the biological instructions that are contained in a strand of DNA. We follow the recent practice of considering DNA sequencing data as text \citep{dotan2024effect, qiao2024model, malusare2024understanding, eapen2025genomic}, albeit we adopt the simple tokenization technique of considering each DNA base in a sequence as a token. This results in DNA sequences being modeled as high-dimensional vectors with a support of four elements per sequence element ($|\chi|=4$). Additional details are in \Cref{app:dna-details}.

\paragraph{Consistency test.} We assess the reliability of \infosedd\ in a low \gls{MI} regime, and consider the ``\textsc{human vs. worm}'' dataset, which is part of the Genomics Benchmarks by \citet{grevsova2023genomic}, a recently proposed suite with eight regulatory element classification tasks. 
    The dataset consists of $10^5$ DNA sequences, paired with 2 class labels (human or worm), with a median length of 200 bases.
We perform a consistency test as follows: for each sample in the dataset, we randomize the label with probability $\rho$ and compute the \gls{MI} between the DNA sequences and the new set of labels. To construct a reference, we can approximate order-of-magnitude and slope of ground-truth \gls{MI} values with the assumption that, given a classifier trained to predict the label given the genome, the classification error is approximately the same across each sample. 
To do so, we express \gls{MI} as $I(X,Y)=H(Y)-H(Y|X)$, where $Y$ is the label and $X$ is the genome, and approximate $H(Y|X)$ as $H_b(\text{Acc.})$, where $H_b$ is the binary entropy and \text{Acc.} is the accuracy of the classifier. 
We estimate $H(Y)$ directly from data using a bin estimate given that we have a binary label and enough samples. In our experiments, we use a pretrained \textsc{caduceus} model \citep{schiff2024caduceus} for the backbone of all methods, with minimal architectural changes when needed, ensuring a fair comparison. For more details check \Cref{app:dna-details}.

\begin{figure}[htbp]
    \centering
    \begin{minipage}[b]{0.45\linewidth}
        \centering
        \includegraphics[scale=0.30]{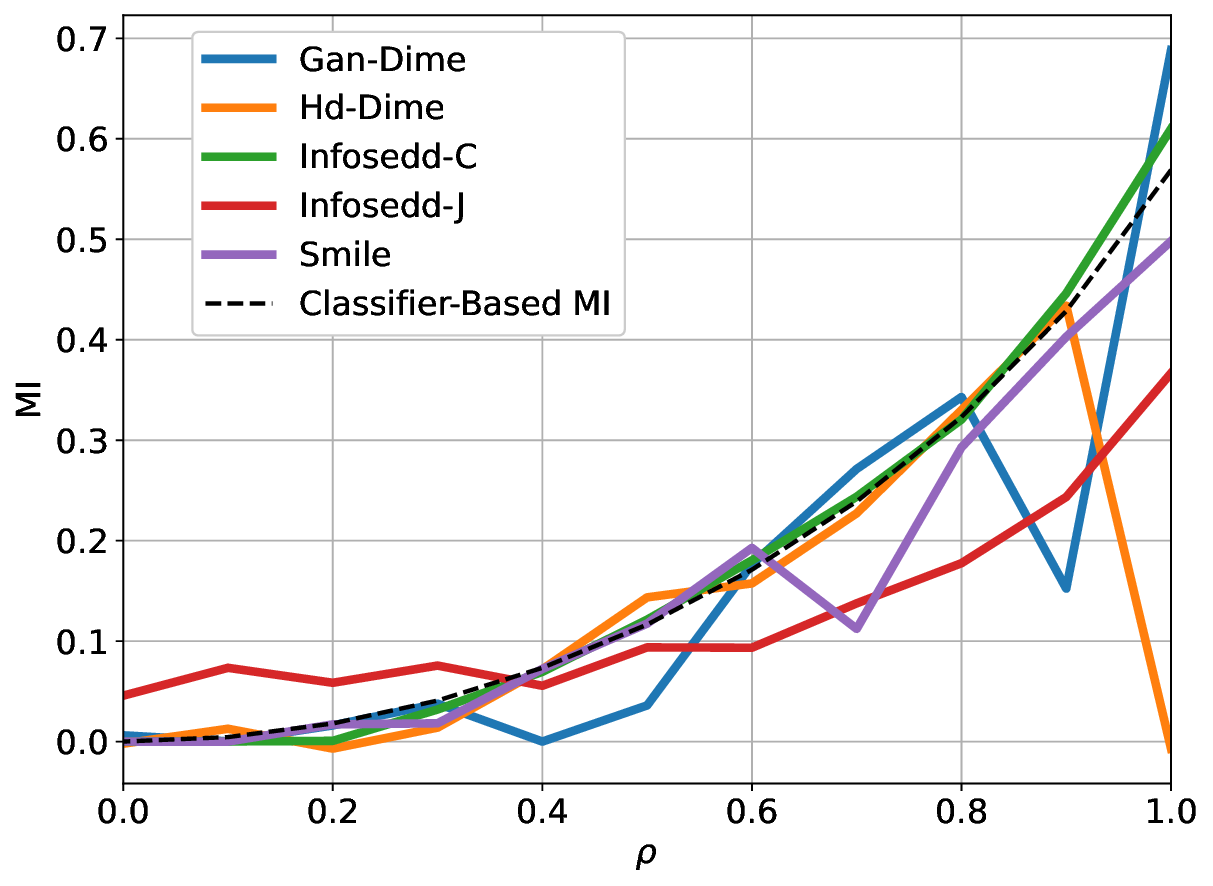}
        \caption{Consistency test for the \textsc{human vs. worm} dataset against the best performing competitors. For the complete numerical results check \Cref{app:dna-details}.}
        \label{fig:consistency-dna}
    \end{minipage}
    \hfill
    \begin{minipage}[b]{0.45\linewidth}
        \centering
        \includegraphics[width=\linewidth]{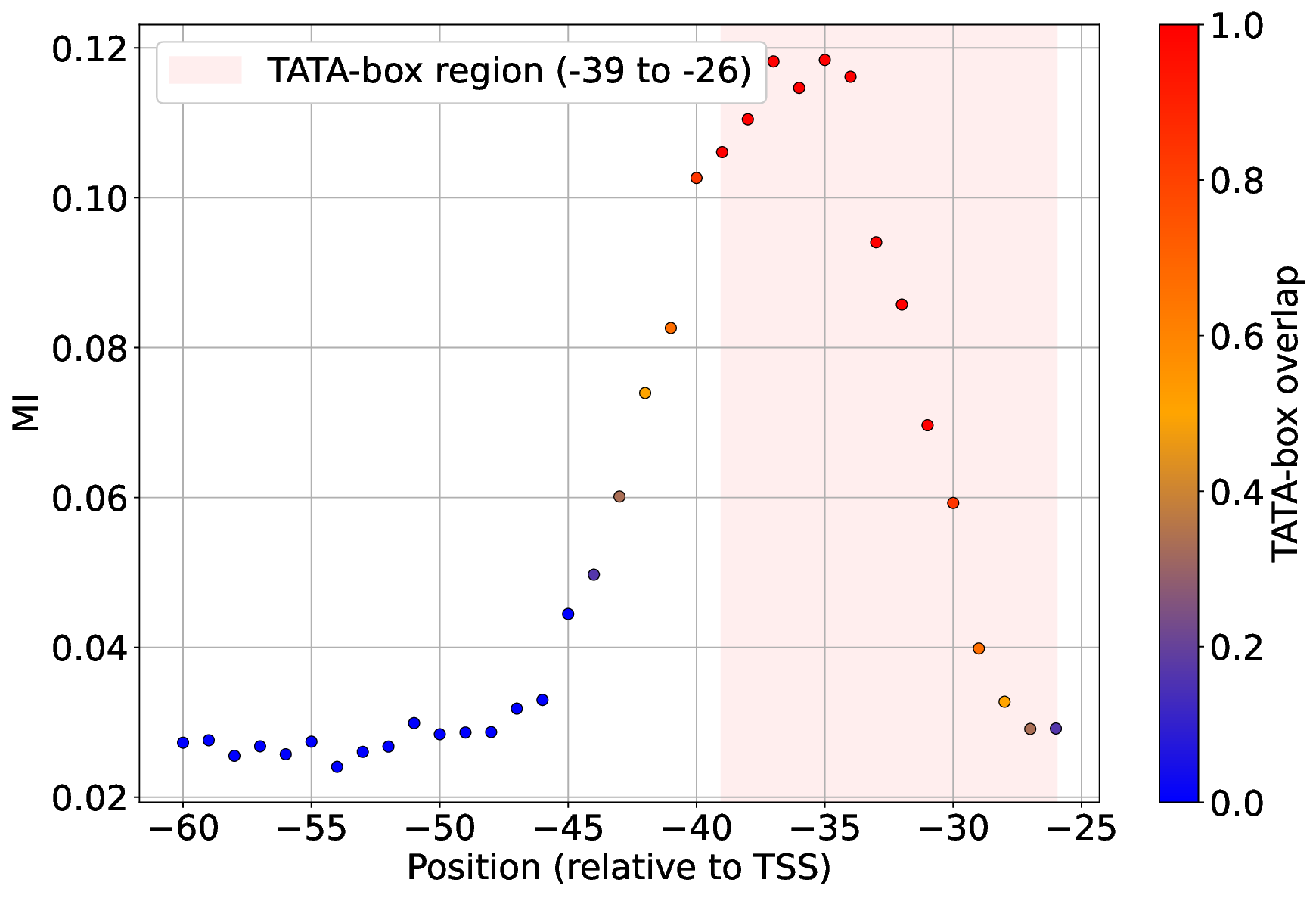}
        \caption{\textit{Arabidopsis thaliana} promoter \textsc{tata-box} identification on DNA sequences using \infosedd.}
        \label{fig:tata-box-search}
    \end{minipage}
\end{figure}

Results in \Cref{fig:consistency-dna} show that \infosedd-C outperforms the competitors, as it closely matches the \textit{classifier-based} reference \gls{MI}. \textsc{hd-dime} performs relatively well for $\rho < 1.0$ while other competitors struggle to keep increasingly high \gls{MI} estimates for growing values of $\rho$. The difference in performance between the two \infosedd\ variants is due to the difference in dimensionality between the label, mono-dimensional, and the DNA sequence. This makes the optimization process of \infosedd-C significantly easier, since it only requires parametrizing the scores of the marginal distribution of the label and of the conditional distribution of the label given the associated DNA sequence. \infosedd-J, instead, must parametrize the score of the joint distribution of the DNA sequence and label, a considerably more difficult task since it needs to do discrete diffusion training on the DNA sequence while \infosedd-C only treats the DNA sequence as a conditioning signal. 

\paragraph{Finding motifs in promoters.} A central issue in molecular biology is understanding the regulatory mechanisms that control gene expression. In particular, gene expression requires a stretch of regulatory DNA called a promoter which contains certain motifs, i.e., patterns that show a statistically significant dependency with expression levels.
Computational methods based on \gls{MI} \citep{elemento2007universal, rao2007motif} use whole genome sequences to search for the key elements in transcription regulation. The key idea is to directly quantify the dependency between the presence or absence of a given motif in a regulatory region and the expression of the corresponding gene. 

\citet{umarov2017recognition} propose a convolutional classifier for building eukaryotic promoter recognition models, which they evaluate on numerous promoter sequences extracted from the well-known \textsc{epd} database \citep{dreos2013epd}. In this work, we focus on \textit{Arabidopsis thaliana} \textsc{tata}-promoter/non-promoter sequences, consisting of 1497 samples and 2879 samples respectively, each of fixed length of 251 DNA bases. 
A minimal eukaryotic promoter region contains a transcription start site (TSS) and a so-called \textsc{tata-box} motif at a position $\approx30$ base pair upstream from the transcription start site, which is a key feature used for classification. 
The \textit{Arabidopsis thaliana} is no stranger to this rule, as the preferred position for the \textsc{tata-box} motif is between -39 and -26 with respect to TSS \citep{bernard2010tc}. 
\citet{umarov2017recognition} propose a randomized technique, which we extend to our method, to find the \textsc{tata-box} motif: to discover such sites we train \infosedd\ such that it can estimate the \gls{MI} between the sequence distribution and the promoter labels. Then, we use a sliding window of length $L$ and mask the DNA sequence outside this window. Using sliding windows moving from the beginning of a functional site sequence, we can build a \gls{MI} profile that reflects the effect of a random sequence. Whenever the window covers a segment that is irrelevant for determining whether a sequence is a promoter or not, \gls{MI} remains low. Instead, unmasking the \textsc{tata-box} motif implies a substantial increase in \gls{MI} values. Differently from the method in \citep{umarov2017recognition}, our approach is robust to correlated motifs: masking only one correlated motif results in high classification accuracy due to the other motif still being unmasked; our method, instead, only unmasks the current motif, which allows assessing its importance without interference from correlated motifs.
For \gls{MI} estimation we fine-tune a pretrained \textsc{caduceus} model, used as the backbone for \infosedd-\textsc{j}. We use the joint variant since its training procedure enables label prediction with DNA sequence masking, while the other variant only models the conditional distribution given unmasked data (more details in \Cref{app:dna-details}).

\Cref{fig:tata-box-search} shows the \gls{MI} profile between the window and the promoter label: red dots imply a high overlap between the window and the \textsc{tata-box} location, as determined by \citet{bernard2010tc}. While known from the genomics literature, \infosedd\ can effectively locate the \textsc{tata-box} using \gls{MI}, which makes it an invaluable tool for motif discovery. Note also that other \gls{MI} estimators would need different training runs for each window, whereas \infosedd\ natively supports \gls{MI} estimation between subsets of DNA sequences. This property unlocks applications of \gls{MI} which were previously hindered by inaccurate estimators and poorly scalable methods.

%% file: sections_iclr/conclusions.tex
The ability to measure the non-linear relation among arbitrary data distributions stands as a cornerstone for scientific discovery, but scalable, and efficient methods to accurately estimate information theoretic metrics, without the need for pre-processing workarounds, has only recently attracted the attention of the community. In this work, we contributed a novel method tailored to estimate such metrics for a particularly challenging case, that of high-dimensional discrete data distributions. To the best of our knowledge, our method is unique, as it is scalable, consistent, and accurate, as demonstrated by our experimental validation; moreover, our approach is efficient, as it can leverage existing pretrained models and does not require ad-hoc procedures to work with discrete data. Remarkably \infosedd\ remains flexible to the choice of masked discrete diffusion model, enabling the usage of other models and parameterizations such as \textsc{md4} \citep{shi2024simplified} or \textsc{LLaDA} \citep{nie2025large}.

Armed with \infosedd, we demonstrated that the ability to estimate \gls{MI} between discrete random variables is key for a number of applications. In the context of text summarization, \gls{MI} is a meaningful proxy to judge performance, and it can be used for model selection. In the genetics domain, we showed that access to accurate \gls{MI} estimates has the potential of enabling new discoveries using sequencing data, which is discrete by nature. Moreover, the solid mathematical foundations of \infosedd\ enable exciting extensions to mixed continuous/discrete data using the Generator Matching framework \citep{holderrieth2024generator}, with many potential applications in diverse scientific fields.




\newpage 

\section*{Reproducibility statement}
We provide the code at \url{https://github.com/AlbertoForesti/mutinfo-diffusion}, where we also provide instructions and utility scripts to build datasets and run experiments. In \Cref{app:model_text} we report the model setup for text experiments and in \Cref{app:model_dna}. Likewise, details and assumptions made in building the datasets are provided in \Cref{app:dataset_synthetic} for synthetic dataset, \Cref{subsubsec:app-text} for text and \Cref{subsubsec:app-genome} for DNA. We do a performance analysis of the benchmarked methods using the same codebase and the same machine, with the only differences being the methodology in \gls{MI} estimation and loss functions.

\section*{Acknowledgements}
Alberto Foresti, Giulio Franzese and Pietro Michiardi were partially funded by project MUSECOM2- AI-enabled MUltimodal SEmantic COMmunications and COMputing, in the Machine Learning based Communication Systems, towards Wireless AI (WAI), Call 2022, ChistERA.

%% file: sections_iclr/appendix.tex
\newpage

\section{Mathematical proofs of the main method}

\subsection{Proof of \eqref{eq:infosedd}} We first break $\expected\left[\log\tfrac{\bp_T}{\bq_T}\bigg|\bproc_0\right]$ in $\expected\left[\log\bp_T\bigg|\bproc_0\right]-\expected\left[\log\bq_T\bigg|\bproc_0\right]$ and apply the Dynkin's formula separately. We start with $\expected\left[\log\bp_T\bigg|\bproc_0\right]$:

{\scriptsize\begin{align}
    \expected\left[\log\bp_T\bigg|\bproc_0\right] \nonumber&= \log\bp_0(\bproc_0)+\expected\left[\int_0^T\tfrac{\partial \log\bp_t}{\partial t}(\bproc_t,t)+\boperator[\log\bp_t](\bproc_t,t)dt\bigg|\bproc_0\right] \\
    & = \log\bp_0(\bproc_0)+\expected\left[\int_0^T\tfrac{\partial \log\bp_t}{\partial t}(\bproc_t,t)+\sum_{x\neq \bproc_t}\bratesp_t(x,\bproc_t)(\log\bp_t(x)-\log\bp_t(\bproc_t))dt\bigg|\bproc_0\right] \nonumber\\
    & = \log\bp_0(\bproc_0)+\expected\left[\int_0^T\tfrac{\partial \log\bp_t}{\partial t}(\bproc_t,t)+\sum_{x\neq \bproc_t}\bratesp_t(x,\bproc_t)\log\tfrac{\bp_t(x)}{\bp_t(\bproc_t)}dt\bigg|\bproc_0\right] \nonumber
\end{align}}

We now focus on the term $\tfrac{\partial \log\bp_t}{\partial t}$, which we can rewrite using $\partial \bp_t=\bratesp_t\bp_t$:

\begin{align*}
    \tfrac{\partial \log\bp_t}{\partial t} = \tfrac{\partial \bp_t}{\partial t}\bigg/ \bp_t = \tfrac{\bratesp_t\bp_t}{\bp_t}
\end{align*}

Where the division between numerator and denominator in the last two terms denotes element-wise division. We focus now on simplifying the numerator, using the definition in \eqref{eq:ratesdef}. Recalling that

\begin{align*}
    \bratesp_t(b,a) &= \left(\tfrac{\fp_{T-t}(b)}{\fp_{T-t}(a)} \fratesp_{T-t}(a,b)\right) (1-\delta(a,b))+\left(-\sum_{b\neq a}\bratesp_t(b,a)\right)\delta(a,a) \\ &= \left(\tfrac{\bp_{t}(b)}{\bp_{t}(b)} \fratesp_{T-t}(a,b)\right) (1-\delta(a,b))+\left(-\sum_{b\neq a}\bratesp_t(b,a)\right)\delta(a,a)
\end{align*}

we compute the $a$-th element of $\bratesp_t\bp_t$:

\begin{align*}
    [\bratesp_t\bp_t](a) &= \sum_b \bratesp_t(a,b)\bp_t(b) \\&= \sum_{b\neq a} \bratesp_t(a,b)\bp_t(b) + \bratesp_t(a,a)\bp_t(a)\nonumber\\
    &= \sum_{b\neq a} \fratesp_{T-t}(b,a)\tfrac{\bp_t(a)}{\bp_t(b)}\bp_t(b) - \sum_{b\neq a}\bratesp_t(b,a)\bp_t(a)\nonumber\\
    &= \sum_{b\neq a} \fratesp_{T-t}(b,a)\bp_t(a) - \sum_{b\neq a}\fratesp_{T-t}(a,b)\tfrac{\bp_t(b)}{\bp_t(a)}\bp_t(a)\nonumber\\
    &= \sum_{b\neq a} \fratesp_{T-t}(b,a)\bp_t(a) - \sum_{b\neq a}\fratesp_{T-t}(a,b)\bp_t(b)\nonumber
    \\
    &= \sum_{b\neq a} \fratesp_{T-t}(b,a)\bp_t(a) - \fratesp_{T-t}(a,b)\bp_t(b)\nonumber
\end{align*}

Finally, if we divide by the denominator we obtain:

{\begin{align*}
    \left[\tfrac{\bratesp_t\bp_t}{\bp_t}\right](a) = &\tfrac{\sum_{b\neq a} \fratesp_{T-t}(b,a)\bp_t(a) - \fratesp_{T-t}(a,b)\bp_t(b)}{\bp_t(a)} \\ = &  \sum_{b\neq a} \fratesp_{T-t}(b,a) - \fratesp_{T-t}(a,b)\tfrac{\bp_t(b)}{\bp_t(a)} \nonumber
\end{align*}}

Moreover, if we notice that $\bratesp_t(x,\bproc_t)\log\tfrac{\bp_t(x)}{\bp_t(\bproc_t)}=\fratesp_{T-t}(\bproc_t,x)\tfrac{\bp_t(x)}{\bp_t(\bproc_t)}\log\tfrac{\bp_t(x)}{\bp_t(\bproc_t)}$, we can write the Dynkin's formula as:

{\begin{align}\label{eq:dynkinp}
    &\expected\left[\log\bp_T|\bproc_0\right] =  \nonumber\\ &\log\bp_0(\bproc_0)+ \nonumber\\ &\expected\left[\int_0^T\sum_{x\neq \bproc_t} \fratesp_{T-t}(x,\bproc_t) - \fratesp_{T-t}(\bproc_t,x)\tfrac{\bp_t(x)}{\bp_t(\bproc_t)}+\fratesp_{T-t}(\bproc_t,x)\tfrac{\bp_t(x)}{\bp_t(\bproc_t)}\log\tfrac{\bp_t(x)}{\bp_t(\bproc_t)}dt\bigg|\bproc_0\right] 
\end{align}}

Then, if we define $K(\alpha)=\alpha(\log \alpha - 1)$ and group some terms we obtain:

\begin{align*}
    \expected\left[\log\bp_T|\bproc_0\right] &= \\ \log\bp_0(\bproc_0)&+\expected\left[\int_0^T\sum_{x\neq \bproc_t} \fratesp_{T-t}(x,\bproc_t) + \fratesp_{T-t}(\bproc_t,x)K\left(\tfrac{\bp_t(x)}{\bp_t(\bproc_t)}\right)dt\bigg|\bproc_0\right] \nonumber
\end{align*}

We now repeat similar calculations for $\expected\left[\log\bq_T\bigg|\bproc_0\right]$. Firstly, the term $\tfrac{\partial \log\bq_t}{\partial t}$ becomes:

\begin{equation*}
    \left[\tfrac{\partial \log\bq_t}{\partial t}\right](a) = \sum_{b\neq a} \fratesp_{T-t}(b,a) - \fratesp_{T-t}(a,b)\tfrac{\bq_t(b)}{\bq_t(a)}
\end{equation*}

Whereas the backward operator term $\boperator[\log\bq_t](\bproc_t,t)$ becomes:
\begin{equation*}
    \bratesp_t(x,\bproc_t)\log\tfrac{\bq_t(x)}{\bq_t(\bproc_t)}=\fratesp_{T-t}(\bproc_t,x)\tfrac{\bp_t(x)}{\bp_t(\bproc_t)}\log\tfrac{\bq_t(x)}{\bq_t(\bproc_t)}
\end{equation*}

Combining everything together gives:

{\begin{align}\label{eq:dynkinq}
    &\expected\left[\log\bq_T\bigg|\bproc_0\right] = \nonumber \\ &\log\bq_0(\bproc_0)+ \nonumber\\&\expected\left[\int_0^T\sum_{x\neq \bproc_t} \fratesp_{T-t}(x,\bproc_t) - \fratesp_{T-t}(\bproc_t,x)\tfrac{\bq_t(x)}{\bq_t(\bproc_t)}+\fratesp_{T-t}(\bproc_t,x)\tfrac{\bp_t(x)}{\bp_t(\bproc_t)}\log\tfrac{\bq_t(x)}{\bq_t(\bproc_t)}dt\bigg|\bproc_0\right] 
\end{align}}

Finally, we can estimate $\expected\left[\log\tfrac{\bp_T}{\bq_T}\bigg|\bproc_0\right]$ by subtracting \eqref{eq:dynkinq} from \eqref{eq:dynkinp}:

{\scriptsize\begin{equation}
    \expected\left[\log\tfrac{\bp_T}{\bq_T}\bigg|\bproc_0\right] \approx \expected\left[\int_0^T\sum_{x\neq \bproc_t} \fratesp_{T-t}(\bproc_t,x)\left(K\left(\tfrac{\bp_t(x)}{\bp_t(\bproc_t)}\right)+\tfrac{\bq_t(x)}{\bq_t(\bproc_t)}-\tfrac{\bp_t(x)}{\bp_t(\bproc_t)}\log\tfrac{\bq_t(x)}{\bq_t(\bproc_t)}\right)dt\bigg|\bproc_0\right] \nonumber
\end{equation}}

By using the fact that $\expected\left[\expected\left[\log\tfrac{\bp_T}{\bq_T}\bigg|\bproc_0\right]\right]=\expected\left[\log\tfrac{\bp_T}{\bq_T}\right]$, $\bp_t=\fp_{T-t}$, $\bq_t=\fq_{T-t}$, $\bproc_t=\fproc_{T-t}$ and by setting $\tau=T-t$, we get:

\begin{align*}
&\expected\left[\int_0^T\sum_{x\neq \fproc_\tau} \fratesp_{\tau}(\fproc_\tau,x)\left(K\left(\tfrac{\fp_\tau(x)}{\fp_\tau(\fproc_\tau)}\right)+\tfrac{\fq_\tau(x)}{\fq_\tau(\fproc_\tau)}-\tfrac{\fp_\tau(x)}{\fp_\tau(\fproc_\tau)}\log\tfrac{\fq_\tau(x)}{\fq_\tau(\fproc_\tau)}\right)d\tau\right]
\end{align*}

recovering \Cref{eq:infosedd}.

\subsection{Entropy estimation details}\label{subsec:absorb_ratio}

We provide hereafter for completeness the steps necessary to show that $\tfrac{\fu_t(x)}{\fu_t(\absorb)}=\ratio$ for $x\neq\absorb$:
{\small\begin{align}
    \tfrac{\fu_t(x)}{\fu_t(\absorb)} = \tfrac{\sum_{x_0\in\support}\fu_t(x|x_0)\fu_0(x_0)}{\sum_{x_0\in\support}\fu_t(\absorb|x_0)\fu_0(x_0)}
    = \tfrac{\sum_{x_0\in\support}\delta(x,x_0)e^{-\cumnoise(t)}\tfrac{1}{N}}{\sum_{x_0\in\support}(1-e^{-\cumnoise(t)})\tfrac{1}{N}} \nonumber = \tfrac{e^{-\cumnoise(t)}}{N(1-e^{-\cumnoise(t)})} = \ratio
\end{align}}

\subsection{Proof of \Cref{eq:free_marginal}}\label{subsec:free_marginal}

Consider $\fp_t(\fproc_t=\xbar,\fprocy_t=\absorb)$:

\begin{align}\label{eq:marg_score}
    &\fp_t(\fproc_t=\xbar,\fprocy_t=\absorb) = \sum_{x,y}\prob(\fproc_t=\xbar, \fprocy_t=\absorb, \fproc_0=x,\fprocy_0=y)\\ &= \sum_{x,y}\underset{\pjump}{\underbrace{\prob(\fprocy_s=\absorb\g\fproc_t=\xbar,\fproc_0=x,\fprocy_0=y)}}\prob(\fproc_t=\xbar\g\fproc_0=x,\fprocy_0=y)\prob(\fproc_0=x,\fprocy_0=y) \nonumber \\ &= \sum_{x,y}\pjump\prob(\fproc_t=\xbar\g\fproc_0=x)\prob(\fproc_0=x,\fprocy_0=y) \nonumber \\ &= \pjump\sum_x\prob(\fproc_t=\xbar\g\fproc_0=x)\underset{\prob(\fproc_0=x)}{\underbrace{\sum_y\prob(\fproc_0=x,\fprocy_0=y)}} \nonumber \\ & = \pjump\prob(\fproc_t=\xbar) \nonumber
\end{align}

\Cref{eq:marg_score} implies that $\tfrac{\fp_t(\fproc_t=x,\fprocy_t=\absorb)}{\fp_t(\fproc_t=\xbar,\fprocy_t=\absorb)} = \tfrac{\prob(\fproc_t=x)}{\prob(\fproc_t=\xbar)}$. This important property enables the estimation of mutual information without modifying the score network.

\section{Algorithm Pseudo-code}\label{app:pseudocode}

In this Section we show the pseudocodes for the different  \infosedd\ variants.

\begin{algorithm}[h]
\caption{\infosedd\textsc{-j}: Estimate $I(\rvx,\rvy)$}
\label{alg:infosedd-j}
\begin{algorithmic}[1]
{\small
    \REQUIRE Initial sample $[\fproc_0,\fprocy_0]\sim \fp_0$, score network $\scorefn$
    \STATE $t\sim u(0,T)$
    \COMMENT{Sample time uniformly}
    \STATE $[\fproc_t,\fprocy_t]\sim \fp_t(\cdot | [\fproc_0,\fprocy_0])$
    \COMMENT{Perturb data}
    \STATE $\hat{I}=0$
    \FOR{$i: \fproc^i_t=\absorb$}
    \FOR{$n\in [1:N]$}
    \STATE {\scriptsize $\tilde{X}=[\fproc^1_t,\dots,\fproc^{i-1}_t,n,\fproc^{i+1}_t,\dots,\fproc^{M}_t]$}
    \STATE {\scriptsize$\hat{I}+=T\sigma(t)\left(K\left(\scorefn([\fproc_t,\fprocy_t])_{[\tilde{X},\fprocy_t]}\right)+\scorefn([\fproc_t,\absorb])_{[\tilde{X},\absorb]}-\scorefn([\fproc_t,\fprocy_t])_{[\tilde{X},\fprocy_t]}\log\left(\scorefn([\fproc_t,\absorb])_{[\tilde{X},\absorb]}\right)\right)$}
    \ENDFOR
    \ENDFOR
    \FOR{$i: \fprocy^i_t=\absorb$}
    \FOR{$n\in [1:N]$}
    \STATE {\scriptsize$\tilde{Y}=[\fprocy^1_t,\dots,\fprocy^{i-1}_t,n,\fprocy^{i+1}_t,\dots,\fprocy^{M}_t]$}
    \STATE {\scriptsize$\hat{I}+=T\sigma(t)\left(K\left(\scorefn([\fproc_t,\fprocy_t])_{[\fproc_t,\tilde{Y}]}\right)+\scorefn([\absorb,\fprocy_t])_{[\absorb,\tilde{Y}]}-\scorefn([\fproc_t,\fprocy_t])_{[\fproc_t,\tilde{Y}]}\log\left(\scorefn([\absorb,\fprocy_t])_{[\absorb,\tilde{Y}]}\right)\right)$}
    \ENDFOR
    \ENDFOR
    \RETURN $\hat{I}$}
    \end{algorithmic}
\end{algorithm}

\begin{algorithm}[h]
\caption{\infosedd-\textsc{c}: Estimate $I(\rvx,\rvy)$}
\label{alg:infosedd-c}
\begin{algorithmic}[1]
{\small
    \REQUIRE Initial sample $[\fproc_0,\fprocy_0]\sim \fp_0$, score network $\scorefn$
    \STATE $t\sim u(0,T)$
    \COMMENT{Sample time uniformly}
    \STATE $\fprocy_t\sim \fp_t(\cdot | \fprocy_0)$
    \COMMENT{Perturb data}
    \STATE $\hat{I}=0$
    \FOR{$i: \fprocy^i_t=\absorb$}
    \FOR{$n\in [1:N]$}
    \STATE {\scriptsize $\tilde{X}=[\fprocy^1_t,\dots,\fprocy^{i-1}_t,n,\fprocy^{i+1}_t,\dots,\fprocy^{M}_t]$}
    \STATE {\scriptsize$\hat{I}+=T\sigma(t)\left(K\left(\scorefn([\fproc_0,\fprocy_t])_{[\fproc_0,\tilde{Y}]}\right)+\scorefn([\absorb,\fprocy_t])_{[\absorb,\tilde{Y}]}-\scorefn([\fproc_t,\fprocy_0])_{[\tilde{X},\fprocy_0]}\log\left(\scorefn([\absorb,\fprocy_t])_{[\absorb,\tilde{Y}]}\right)\right)$}
    \ENDFOR
    \ENDFOR
    \RETURN $\hat{I}$}
    \end{algorithmic}
\end{algorithm}

\begin{algorithm}[H]
\caption{\textsc{info-sedd-h}: Estimate $H(\rvx)$}
\label{alg:compute_entropy}
\begin{algorithmic}[1]
{\small
  \REQUIRE Initial sample $\fproc_0\sim \fp_0$, score network $\scorefn$
    \STATE $t\sim u(0,T)$
    \COMMENT{Sample time uniformly}
    \STATE $\fproc_t\sim \fp_t(\cdot | \fproc_0)$
    \COMMENT{Perturb data}
    \STATE $\hat{H}=0$
    \FOR{$i: \fproc^i_t=\absorb$}
    \FOR{$n\in [1:N]$}
    \STATE $\tilde{X}=[\fproc^1_t,\dots,\fproc^{i-1}_t,n,\fproc^{i+1}_t,\dots,\fproc^{M}_t]$
    \STATE $\hat{H}+=T\sigma(t)\left(K\left(\scorefn(\fproc_t)_{\tilde{X}}\right)+\ratio-\scorefn(\fproc_t)_{\tilde{X}}\log\left(\ratio\right)\right)$
    \ENDFOR
    \ENDFOR
    \RETURN $\hat{H}$
}
    \end{algorithmic}
\end{algorithm}

\section{Experimental details}

\subsection{Synthetic experiments}\label{app:synthetic}

\subsubsection{Dataset construction}\label{app:dataset_synthetic}
In this section, we show how we construct the datasets for our synthetic experiments. We generate joint distributions for random variables $X, Y$ with user-defined mutual information and support sizes $\support_X, \support_Y$. Using an evolutionary strategy, we encode the joint distribution in a vector $g_a \in \mathbb{R}^{|\support_X||\support_Y|}$ and transform it into a valid probability distribution via normalization and reshaping: $\tfrac{g_a-\text{min}(g_a)\ones+\epsilon\ones}{\ones(g_a-\text{min}(g_a)\ones+\epsilon\ones)}$ where $\epsilon$ ensures full support. The mutual information, computable in closed form, serves as the selection criterion in the evolutionary process. For large $|\support_X|$ and $|\support_Y|$, the evolutionary strategy struggles. Instead, we generate high mutual information distributions by concatenating independent distributions, leveraging the additive property of mutual information. Additionally, isomorphisms like Cantor’s pairing function, $\pi(x,y) = \tfrac{1}{2}(x+y)(x+y+1)+y$, enables support expansion without altering mutual information, aiding consistency across experiments. \\
More formally, we exploit the additivity of mutual information with independent random variables to generate complex datasets. By appending discrete noise random variables $Z_{\rvx}$, $Z_{\rvy}$ to the original random variables $\rvx$, $\rvy$, we have $I(\rvx,\rvy) = I([\rvx, Z_{\rvx}],[\rvy, Z_{\rvy}])$. Pairing functions are isomorphisms that map $\sN\times\sN$ to $\sN$. They allow preserving mutual information through the Markov Chain:

\begin{equation*}
    [\rvx,Z_{\rvx}] \to \hat{\rvx} \to \hat{\rvy} \to [\rvy,Z_{\rvy}] \to \hat{\rvy} \to \rvx \to [\rvx,Z_{\rvx}]
\end{equation*}

Where, by the \textit{data processing inequality}, we have $I([\rvx,Z_{\rvx}],[\rvy,Z_{\rvy}]) \geq I(\hat{\rvx},\hat{\rvy}) \geq I([\rvx,Z_{\rvx}],[\rvy,Z_{\rvy}]) \implies I([\rvx,Z_{\rvx}],[\rvy,Z_{\rvy}]) = I(\hat{\rvx},\hat{\rvy}) \implies I(\hat{\rvx},\hat{\rvy}) = I(\rvx,\rvy)$.

In our experiments, we keep the same support dimension for both random variables, increasing it using the following procedure:
\begin{enumerate}
    \item We sample two binomial random variables $Z_{\rvx},Z_{\rvy}$ with parameters $(n,p)$. 
    We vary $n$ for increasing the complexity of the experiment and we keep $p$ fixed to $0.5$.
    \item We concatenate $Z_{\rvx}$ and $Z_{\rvy}$ respectively to $\rvx$ and $\rvy$, to form higher support versions $\hat{\rvx},\hat{\rvy}$.
    \item We map the noisy $\hat{\rvx},\hat{\rvy}$ versions to univariate random variables by applying Cantor's mapping.
\end{enumerate}

\subsubsection{Model and training setup}\label{subsubsec:training_details_synthetic} In this section, we provide additional experimental details. For all the methods present in the benchmark, we use a Multi Layer Perceptron (MLP) with skip connections and around $130$k parameters, based on the architecture used in \cite{franzese2023minde}, reworking the initial layer to include absolute positional embeddings. 
For training our discrete score model, we match the methodology used by \cite{lou2024discrete}, using the absorb configuration. At inference time, we always take the last valid validation step estimate of each method to avoid not-a-number values in our tables.

\subsubsection{Study on training convergence}\label{app:convergence}
\begin{center}
    \begin{figure}
    \centering
    \includegraphics[width=1.0\linewidth]{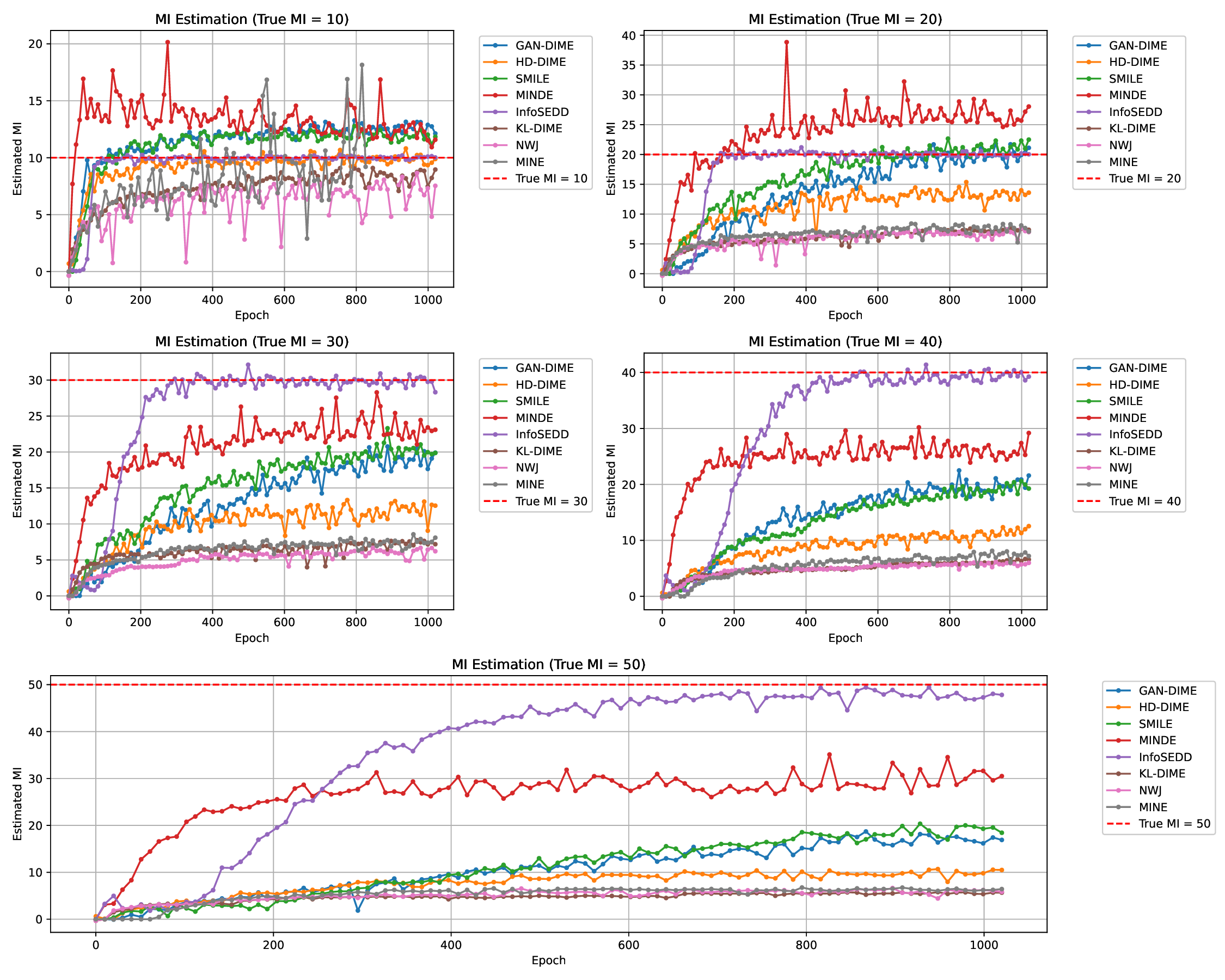}
    \caption{Empirical convergence of the benchmarked \gls{MI} estimators.}
    \label{fig:convergence}
\end{figure}
\end{center}

Since all models were trained in the same conditions, we study the convergence to the correct \gls{MI} for the benchmarked methods. As we can see from \Cref{fig:convergence}, \infosedd\ is the method which converges faster. Other methods, even when estimating the right \gls{MI}, take more epochs to settle on the correct value. We also benchmark runtime and maximum memory consumption for the tasks presented in \Cref{tab:mi_comparison_std}. We run our experiments on a NVIDIA A100-SXM4-80GB GPU and we observed that \infosedd\ outperforms all competitors on memory consumption (\Cref{tab:memory}) and all variational competitors on runtime (\Cref{tab:runtime}).

\input{tables/memory_table}

\input{tables/runtime_table}

\subsubsection{Low dimensional benchmark}
We also benchmark all methods on an easier low-dimensional benchmark. We use the same training setup, except that we train only for $10^4$ steps using $10^4$ samples and use the same backbone but with 8k parameters. As we can see from {\Cref{tab:mi_comparison_low_dim}}, no clear winner emerges, as all estimators besides \textsc{minde} perform a good job in predicting the right \gls{MI}.

\input{tables/synthetic/low_dim}

\subsubsection{Sample complexity benchmark}\label{app:sample_complexity}

The sample complexity of \infosedd\  strictly depends on the sample complexity of the score network. Theoretical guarantees on score networks and discrete diffusion models require a careful and involved analysis, which we defer to future work. Nevertheless, we run an ablation study on sample complexity, from an empirical standpoint. 

We consider a pair of random vectors $X,Y$, both of length $20$. Each component of $X,Y$, in turn, has a support of $4$. We set the ground truth MI to $10,20,30,40$ and $50$ and use an MLP score network with skip connections and $160$k parameters, and train it for $100$k steps. We iterate over the number of available training samples $N$ as reported in the following table.

\begin{table}[h!]
\centering
\caption{Empirical evaluation of sample complexity for \infosedd.}
\begin{tabular}{lcccc}
\toprule
 N & $10^2$ & \textbf{$10^{3}$} & \textbf{$10^{4}$} & \textbf{$10^{5}$} \\
\midrule
MI=10, D=10 & 7.472 & 11.139 & 10.100 & 10.115 \\
MI=20, D=20 & 7.772 & 22.533 & 20.728 & 19.922 \\
MI=30, D=30 & 26.126 & 29.458 & 36.093 & 29.809 \\
MI=40, D=40 & 28.517 & 38.512 & 35.450 & 38.758 \\
MI=50, D=50 & 27.782 & 50.300 & 43.750 & 48.822 \\
\bottomrule
\end{tabular}
\end{table}


\infosedd\  is sufficiently accurate also for a rather low number of samples. However, as most neural estimators, it relies on deep architectures that require a sufficient number of samples for training. Practitioners are required to find the sweet spot between score network architecture complexity (which caters better MI estimates for complex data) and the need for a large number of training samples.

\subsubsection{Ablation study on support dimension}\label{app:big_supp}

In the main paper we performed synthetic experiments by fixing $D$ to $10$, \gls{MI} to 0.5 and increasing $|\chi|$. In this section, we perform the contrary and we increase $\chi$ by using Cantor's mapping and uncorrelated binomial random variables as noise in the data. We increase $\chi|$ as in \Cref{tab:big_supp}, reporting the new support size $\chi$ after the injectio of the noisy binomial random variables. \Cref{tab:big_supp} shows how \infosedd\ is the only estimator robust to increasingly high support size.

\input{tables/synthetic/big_supp}

\subsubsection{Extreme dimensionality experiments}\label{sec:extremedim}

In the main text, we evaluated \infosedd\ on synthetic benchmarks with increasing dimensionality $D$ and mutual information (MI).  
Here, we further assess scalability in very high-dimensional regimes and additionally compare against \textsc{gan-dime}, \textsc{hd-dime}, and \textsc{smile} for the case where the ground-truth MI is $0.5$.

\paragraph{\normalfont Case $D=256$ and \gls{MI}=0.5}

We trained \infosedd\ using the same architecture as in \cref{sec:synthetic} with an extended budget of $10^6$ optimization steps.  
Table~\ref{tab:extreme_d256} reports the MI estimates for all methods.

\begin{table}[h]
\centering
\caption{Extreme dimensionality experiment with $D=256$ and ground-truth MI $=0.5$.}
\label{tab:extreme_d256}
\begin{tabular}{l c}
\toprule
Method & Estimated MI \\
\midrule
\infosedd & 0.43 \\
\textsc{hd-dime} & 0.43 \\
\textsc{smile} & 0.67 \\
\textsc{gan-dime} & 0.51 \\
\bottomrule
\end{tabular}
\end{table}

While competing methods remain relatively stable in this regime, \infosedd\ provides a competitive and well-calibrated estimate.

\paragraph{\normalfont Case $D=1024$ and \gls{MI}=0.5}

For $D=1024$, we adopted a DiMamba backbone with $2$M parameters, trained for $10^6$ steps with early stopping on a dataset of $10^6$ samples.  
Results are reported in Table~\ref{tab:extreme_d1024}.

\begin{table}[h]
\centering
\caption{Extreme dimensionality experiment with $D=1024$ and ground-truth MI $=0.5$.}
\label{tab:extreme_d1024}
\begin{tabular}{l c}
\toprule
Method & Estimated MI \\
\midrule
\infosedd & 0.53 \\
\textsc{hd-dime} & 0.00 \\
\textsc{smile} & 0.00 \\
\textsc{gan-dime} & 0.00 \\
\bottomrule
\end{tabular}
\end{table}

In this extreme regime, all competing methods collapse to zero, whereas \infosedd\ remains accurate and close to the ground-truth value.  
These results provide further evidence of the robustness and scalability of our approach in very high-dimensional settings.

%
%
%
\paragraph{Discussion}  
These results indicate that \infosedd\ can be effectively scaled to very high-dimensional discrete distributions when trained with sufficient data and capacity. 
Together with the experiments in the main paper, they highlight both the robustness of our approach and its ability to leverage larger training budgets when available.

\newpage

\subsection{Text summarization experiments}\label{app:summarization}
\subsubsection{Dataset and ground truth construction}\label{subsubsec:app-text}
We consider random variable $X$, which correspond to source texts as in \cite{fabbri2021summeval}, and summaries $Y$ obtained with the BART model (M22 of the SummEval paper) when fed with the corresponding source text. Then, for a given $0<\rho<1$, we construct $Y^\rho$ which is equal to the original $Y$ with probability $\rho$ and equal to an independent sample from the same distribution of $Y$ with probability $1-\rho$.
Although we cannot establish the ground truth \gls{MI} $I(X,Y^\rho) $ exactly, we can bound its value. Consider the latent variable $Z$, Bernoulli distributed with parameter $\rho$, which determines whether $Y^\rho$ is equal to the original $Y$ or and independent sample from the same distribution. Then

\begin{align*}
    I(X,Y^\rho) &= I(X;Y^\rho|Z)-I(X;Z|Y^\rho)+I(X,Z) \\ &= I(X;Y^\rho|Z)-I(X;Z|Y^\rho) \\ & \geq I(X;Y^\rho|Z) - H(Z|Y^\rho)\\ & \geq I(X;Y^\rho|Z) - H(Z) \\
    &= \rho I(X;Y^\rho|Z=1) +(1-\rho) I(X;Y^\rho|Z=0)- H(Z) \\ &=\rho I(X;Y^\rho|Z=1)- H(Z) \\ &= \rho I(X;Y)- H(Z)
\end{align*}

Furthermore, since $ I(X,Y^\rho) = I(X;Y^\rho|Z)-I(X;Z|Y^\rho) \leq I(X;Y^\rho|Z)=\rho I(X;Y)$, we can claim that  
$I_\rho(X,Y)\in[\rho I(X,Y) - H(Z),\rho I(X,Y)]$. Since $H(Z)=-\left(\rho\log\rho+(1-\rho)\log(1-\rho)\right)$, whenever $I(X;Y)$ is much greater than $H(Z)$, $I(X;Y^\rho)$ satisfies approximately the linear relationship $I(X;Y^\rho|Z)=\rho I(X,Y)$. This assumption is supported by the literature \citep{takahira2016entropy}.

\subsubsection{Model setup}\label{app:model_text}
For all our methods, we use the MDLM-small model \citep{sahoo2025simple} trained on OpenWebText. While this family of models does not directly parametrize the score, \citet{sahoo2025simple} show that it can be easily recovered from the output of the model. For the experiments with MINDE, we use the variance preserving \gls{SDE} and the same model. We perform the diffusion step on the token embeddings. We trained all models using the configuration in \Cref{tab:training_settings0}. 

\begin{table}[H]
\centering
\caption{Training settings for one run for the model selection experiments and the consistency tests (one value of $\rho)$}
\label{tab:training_settings0}
\begin{tabular}{ll}
\toprule
\textbf{Parameter} & \textbf{Value} \\
\midrule
Backbone & MDLM small \citep{sahoo2025simple} \\
Optimizer & AdamW \\
AdamW - $\beta_1$ & 0.9 \\
AdamW - $\beta_2$ & 0.999 \\
Weight decay & 0 \\
Learning rate & 3e-4 \\
Batch size & 64 \\
Number of steps & 3600 \\
Dropout rate & 0.0 \\
Hardware & $1\times$NVIDIA A 100 80 GB\\
\bottomrule
\end{tabular}
\end{table}

\subsubsection{Numerical results}
In \Cref{tab:num_results_consistency_summ}, we report the numerical results for \gls{MI} estimations in the consistency tests, including \textsc{minde}. We report all values in nats.

\begin{table}[ht]
\centering
\caption{Numerical results for Text summarization consistency tests}
\rotatebox{90}{
\begin{tabular}{lrrrrrrrrr}
\toprule
 & \textsc{gan-dime} & \textsc{hd-dime} & \infosedd\textsc{-c} & \infosedd\textsc{-j} & \textsc{kl-dime} & \textsc{minde} & \textsc{mine} & \textsc{nwj} & \textsc{smile} \\
$\rho$ &  &  &  &  &  &  &  &  &  \\
\midrule
0.0 &  -  & 0.09 & 7.52 & 106.69 & -0.01 & 365679296.00 & -0.50 & -0.01 & 0.00 \\
0.1 &  -  & 0.20 & 34.02 & 157.90 & 0.01 & 198489552.00 & -0.02 & 0.00 & 0.00 \\
0.2 &  -  & -0.01 & 61.25 & 197.18 & 0.48 & 347922112.00 & 0.02 & -0.10 & 0.09 \\
0.3 &  -  & 0.13 & 90.17 & 220.47 & 3.03 & 12828973.00 & 0.00 & -0.24 & 0.89 \\
0.4 &  -  & 2.41 & 124.48 & 244.53 & 3.21 & 33884712.00 & 0.84 &  -  & 2.06 \\
0.5 &  -  & 10.57 & 170.03 & 255.42 & 4.63 & 19269294.00 & 0.33 & 0.19 & 5.84 \\
0.6 &  -  & 16.17 & 230.41 & 276.28 & 5.26 & 270945856.00 & -4.11 & 0.84 & 7.19 \\
0.7 & 9.63 & 12.94 & 308.30 & 303.20 & 5.17 & 391443968.00 & -3.68 &  -  & 12.46 \\
0.8 & 9.28 & 28.98 & 409.26 & 316.01 & 6.38 & 448679872.00 & -3.90 &  -  & 13.85 \\
0.9 & 9.33 & 27.12 & 505.13 & 330.76 & 7.95 & 18818858.00 & -3.96 &  -  & 14.73 \\
1.0 & 19.44 & 39.42 & 602.77 & 361.23 & 6.82 & 124222336.00 & -3.96 &  -  & 20.55 \\
\bottomrule
\end{tabular}
}
\label{tab:num_results_consistency_summ}
\end{table}

\newpage

\subsubsection{Detailed results for human metrics correlation experiments}\label{app:mi-human-scatter}

\begin{figure}[H]
    \centering
    \includegraphics[width=0.6\linewidth]{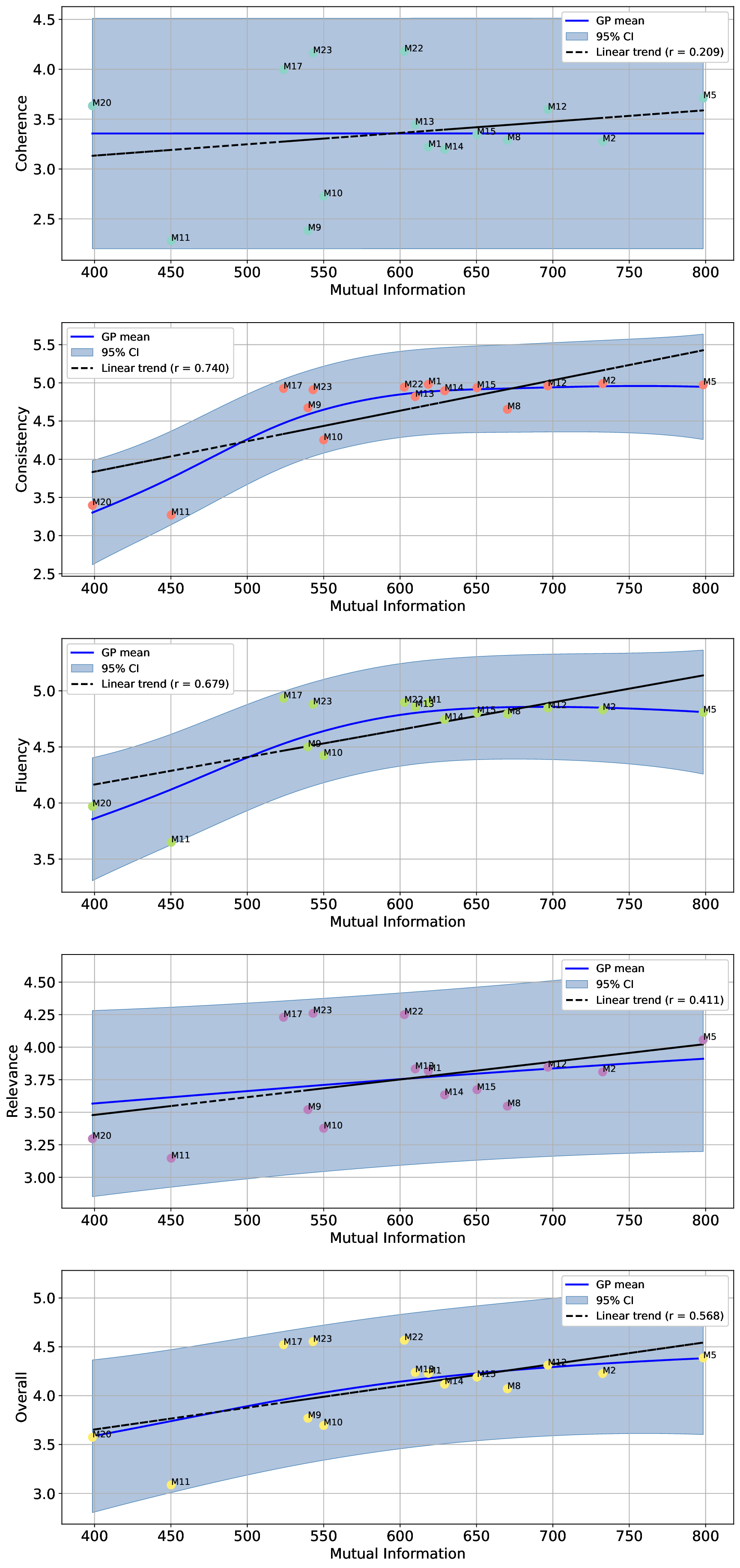}
    \caption{Comparison of human metrics against mutual information estimated by \infosedd-\textsc{c}}
    \label{fig:mi-hm-app-infosedd-c}
\end{figure}

\begin{figure}[H]
    \centering
    \includegraphics[width=0.6\linewidth]{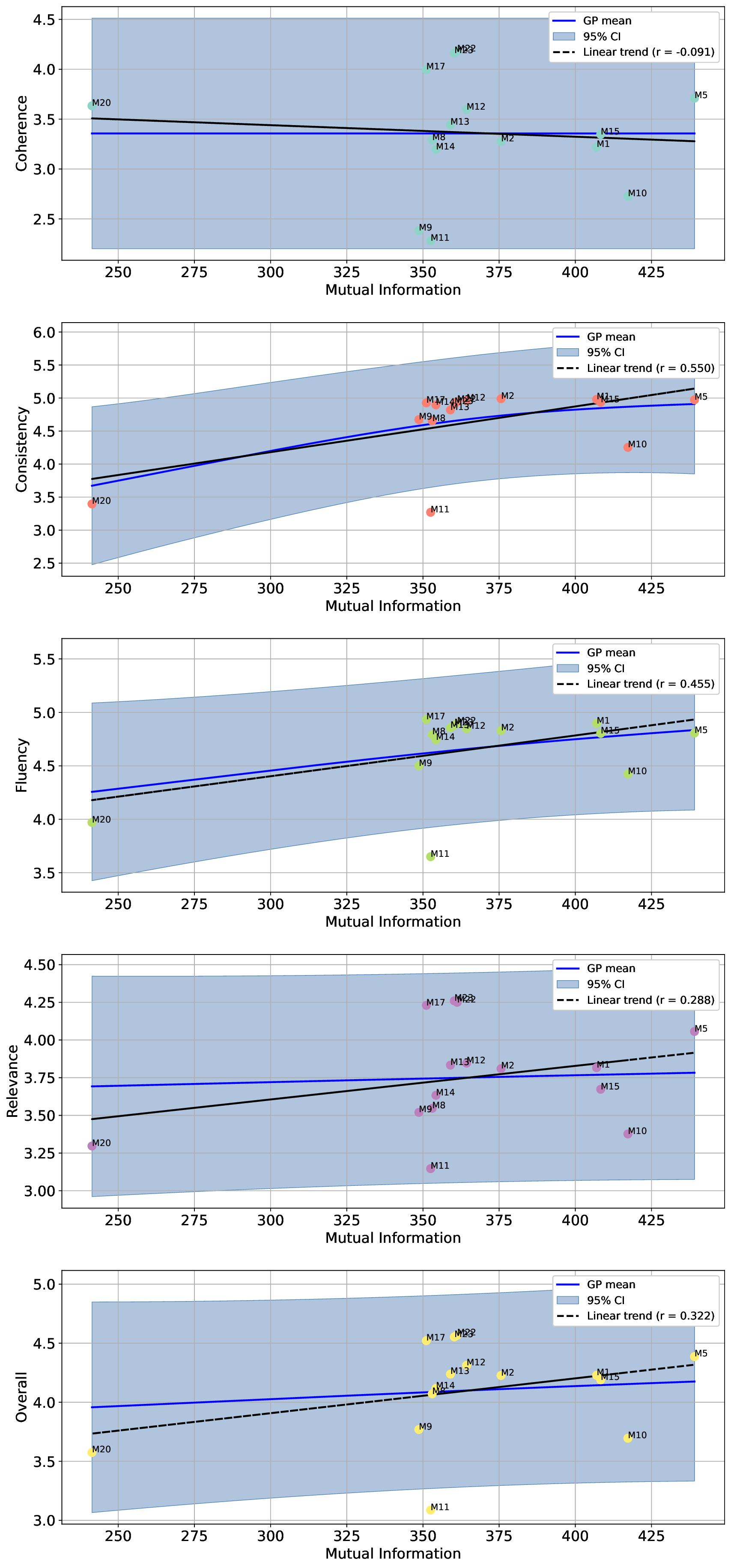}
    \caption{Comparison of human metrics against mutual information estimated by \infosedd-J}
    \label{fig:mi-hm-app-infosedd-j}
\end{figure}

We report the \gls{MI} estimates of each model and compare them with human evaluation metrics. For each metric, we run a \gls{GP} regression with a Matérn kernel with smoothing parameter $\nu=1.5$ \citep{rasmussen2006gaussian} and noise level $0.01$ using the \gls{MI} estimates from both \infosedd-\textsc{c} and \infosedd-\textsc{j}. We use the \textit{scikit-learn} Python package to run this analysis \citep{scikit-learn}. In \Cref{fig:mi-hm-app-infosedd-c,fig:mi-hm-app-infosedd-j}, we plot the GP mean, its 95\% confidence interval, and the Pearson correlation. As the figures show, several human metrics saturate around $5.0$ (the maximum score possible), while \gls{MI} is not subject to this ceiling effect.

In particular, for the consistency and fluency metrics, the GP mean follows an increasing trend until the human scores saturate, suggesting that the linear relationship between \gls{MI} and the human metric only holds up to a threshold, beyond which most models achieve the maximum score. 

Unlike human metrics, however, \infosedd\ does not rely on annotations: it only requires a dataset of summaries paired with source texts, making it a cost-effective and scalable alternative for model selection in text summarization.

\subsection{Comparison of \gls{MI} estimates with \infosedd\ and other text summarization metrics}\label{app:corr_mats}
\begin{figure}[!ht]
    \centering
    \includegraphics[width=0.8\linewidth]{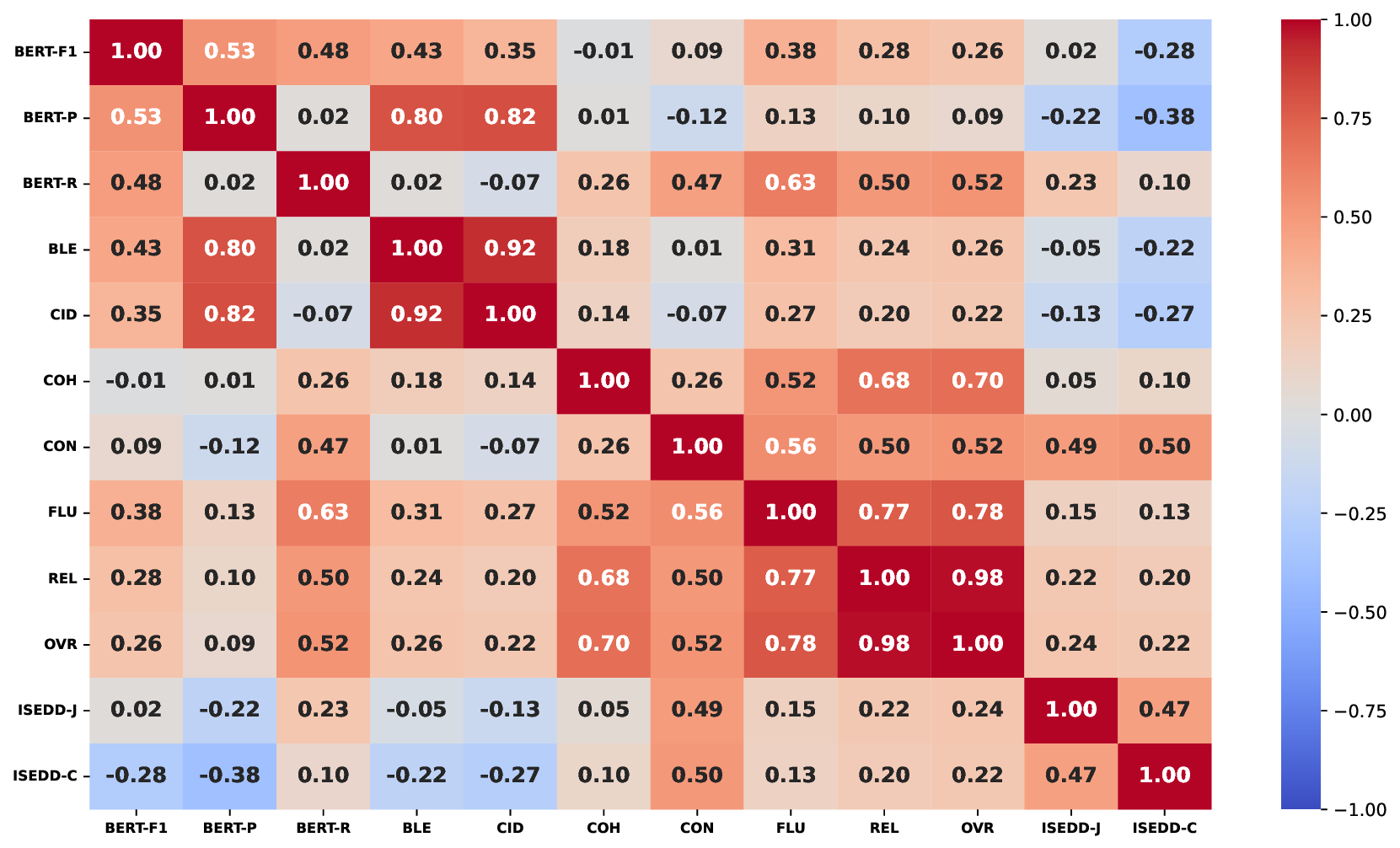}
    \caption{Kendall Tau correlation heatmap of \gls{MI} estimates \infosedd, other text summarization metrics and human metrics. CID=Cider, BLE=bleu.}
    \label{fig:kendall_corrs_all_metrics}
\end{figure}

\begin{figure}[!ht]
    \centering
    \includegraphics[width=0.8\linewidth]{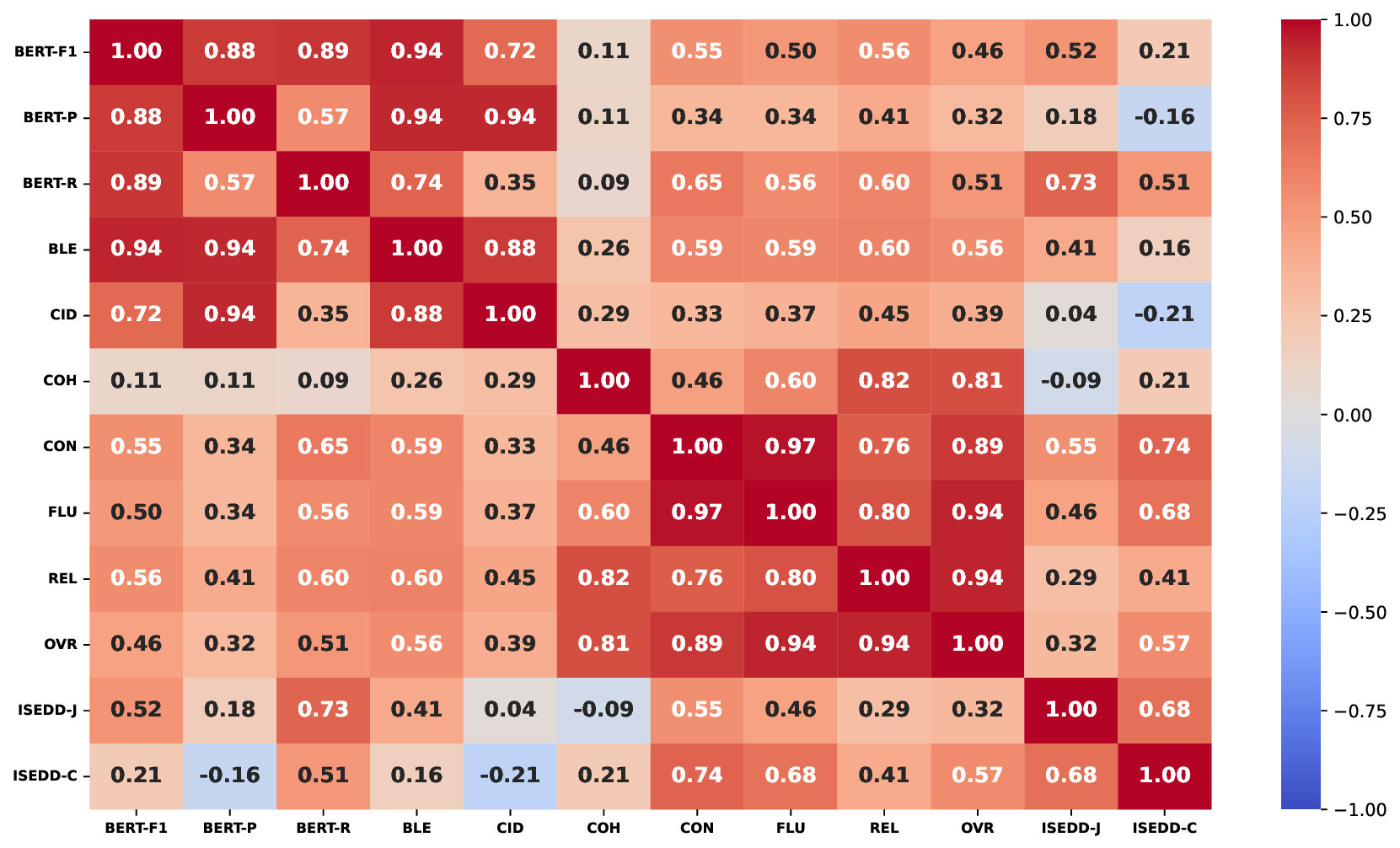}
    \caption{Pearson correlation heatmap of \gls{MI} estimates \infosedd, other text summarization metrics and human metrics. CID=Cider, BLE=bleu.}
    \label{fig:pearson_corrs_all_metrics}
\end{figure}

We measured several other metrics using the SummEval library \citep{fabbri2021summeval} and one reference summary for each sample provided in the SummEval collection. In particular, we benchmark the \textsc{bert score} metric \citep{zhang2019bertscore}, the \textsc{bleu} metric \citep{papineni2002bleu} and the \textsc{cider} metric \citep{vedantam2015cider}. As we can see from \Cref{fig:pearson_corrs_all_metrics} and \Cref{fig:kendall_corrs_all_metrics}, we find that \infosedd-J in general correlates better with other summarization metrics with respect to \infosedd-C. Interestingly, \infosedd-C correlates better with human metrics, proving to be a useful proxy for human metrics without the need of human annotations. Notably, \infosedd-C is also the best at correlating with consistency among all other text summarization metrics.

\newpage

\subsection{Genomics experiments}\label{app:dna-details}
\subsubsection{Dataset and ground truth construction}\label{subsubsec:app-genome}

To construct our dataset for the consistency test, we consider the shuffling hyperparameter $\rho$. We consider $X$ to be the random variable representing the DNA sequences and, for a given $0<\rho<1$, we construct $Y^\rho$ which is equal to the original $Y$, the random variable representing the human or worm label, with probability $1-\rho$ and equal to an independent sample from the same distribution of $Y$ with probability $\rho$.  We can estimate $I(X,Y^\rho)$ by considering the entropy terms which correspond to the decomposition $H(Y^\rho)-H(Y^\rho|X)$. On the one hand, estimation of $H(Y^\rho)=H(Y)$ from the dataset is relatively simple and stable given the low dimensionality of the label. Concerning $H(Y^\rho|X)$, we first write the conditional distribution $p_{Y^\rho|X}(y^\rho|x)$ as:
\begin{align*}
    p_{Y^\rho|X}(y^\rho|x) &= \sum_yp_{Y^\rho,Y|X}(y^\rho,y|x) \\& = \sum_yp_{Y^\rho|Y,X}(y^\rho|y,x)p_{Y|X}(y|x)\\ & = \sum_y(1-\rho)\delta(y,y^\rho)p_{Y|X}(y|x)+\rho p_Y(y^\rho)p_{Y|X}(y|x) \\ &= (1-\rho)p_{Y|X}(y^\rho|x) +\rho p_Y(y^\rho)
\end{align*}

In the considered dataset, classes are balanced and consequently $p_Y(y^\rho)=0.5$. To approximate $p_{Y|X}(y^\rho|x)$, we first define $l(x)=\arg\max_y p_{Y|X}(y|x)$, then we assume the following:

\begin{equation*}
p_{Y|X}(y|x)=
    \begin{cases}
    1-\epsilon &\text{ for } y=l(x)\\
    \epsilon &\text{otherwise}
    \end{cases}
\end{equation*}

with $\epsilon \ll 1$, which roughly speaking corresponds to assuming that for all values of $x$ the conditional probability is highly skewed either towards zero or one. This assumption proves to be reasonable in scenarios where the classification accuracy is close to one, as the case considered here \citep{schiff2024caduceus}. Then

\begin{equation*}
p_{Y^\rho|X}(y^\rho|x)=
    \begin{cases}
    (1-\rho)(1-\epsilon)+\rho p_Y(y^\rho) &\text{ for } y=l(x)\\
    (1-\rho)\epsilon+\rho p_Y(y^\rho) &\text{otherwise}
    \end{cases}
\end{equation*}

Going back to $H(Y^\rho|X)$, we can finally write:
\begin{align*}
    H(Y^\rho|X) &= -\sum_xp_X(x)\sum_y p_{Y^\rho|X}(y^\rho|x)\log p_{Y^\rho|X}(y^\rho|x) \\
    &= -\sum_xp_X(x)( p_{Y^\rho|X}(l(x)|x)\log p_{Y^\rho|X}(l(x)|x) + \\ &+ p_{Y^\rho|X}(1-l(x)|x)\log p_{Y^\rho|X}(1-l(x)|x))\\  &= \sum_xp_X(x)((1-\rho)(1-\epsilon)+\rho p_Y(y^\rho))\log((1-\rho)(1-\epsilon)+\rho p_Y(y^\rho)) \\ &+ \sum_xp_X(x)((1-\rho)\epsilon+\rho p_Y(y^\rho))\log((1-\rho)\epsilon+\rho p_Y(y^\rho)) \\ &= \sum_xp_X(x)H_b((1-\rho)(1-\epsilon)+\rho 0.5) = H_b((1-\rho)(1-\epsilon)+\rho 0.5)
\end{align*}

To select $\epsilon$, we choose the accuracy of the best classifier from \citep{schiff2024caduceus} trained for the \textsc{human vs worm} classification task.

\subsubsection{Model setup}\label{app:model_dna}
We employed a pretrained Caduceus model \citep{schiff2024caduceus} as a backbone for all estimators. We modified the head of the model when needed. For \textsc{mine}-like estimators we added an attention head and added a fully connected layer for aggregating the output. For \textsc{minde}, we added a transformer block with time conditioning. For \infosedd, we did not make architectural variations and we employ the optimized training strategy highlighted in \Cref{app:synthetic}. 
We trained all models using the configuration in \Cref{tab:training_settings}. For the motif selection experiments, we employ \textsc{info-sedd-j} because our method requires masking elements of the DNA sequence, which is something that \textsc{info-sedd-c} does not do during training.

\begin{table}[ht]
\centering
\caption{Training settings for one run for the promoter experiments or the consistency tests (one value of $\rho)$}
\label{tab:training_settings}
\begin{tabular}{ll}
\toprule
\textbf{Parameter} & \textbf{Value} \\
\midrule
Backbone & Caduceus-1K \citep{schiff2024caduceus} \\
Optimizer & AdamW \\
AdamW - $\beta_1$ & 0.9 \\
AdamW - $\beta_2$ & 0.999 \\
Weight decay & 0.1 \\
Learning rate & 0.001 \\
Batch size & 128 \\
Number of steps & 20000 \\
Dropout rate & 0.1 \\
Hardware & $1\times$NVIDIA A 100 80 GB\\
\bottomrule
\end{tabular}
\end{table}

\subsubsection{Numerical results for the consistency tests}
In \Cref{tab:consistency_numerical},  we report the numerical results for \gls{MI} estimations in the consistency tests, including \textsc{minde}. We report all values in nats.

\begin{table}[ht]
\small
\centering
\caption{Numerical results for DNA consistency tests}
\begin{tabular}{lrrrrrrrr}
\toprule
 & \textsc{gan-dime} & \textsc{hd-dime} & \infosedd\textsc{-c} & \infosedd\textsc{-j} & \textsc{kl-dime} & \textsc{mine} & \textsc{nwj} & \textsc{smile} \\
$\rho$ &  &  &  &  &  &  &  &  \\
\midrule
0.0 & 0.01 & 0.00 & 0.00 & 0.05 & 0.00 & 0.00 & -0.02 & 0.00 \\
0.1 & 0.00 & 0.01 & 0.00 & 0.07 & 0.00 & 0.00 & -0.01 & 0.00 \\
0.2 & 0.02 & -0.01 & 0.00 & 0.06 & 0.00 & 0.00 & -0.02 & 0.02 \\
0.3 & 0.04 & 0.01 & 0.03 & 0.08 & 0.00 & 0.00 & -0.01 & 0.02 \\
0.4 & 0.00 & 0.07 & 0.07 & 0.06 & -0.01 & 0.00 & 0.00 & 0.07 \\
0.5 & 0.04 & 0.14 & 0.12 & 0.09 & 0.12 & 0.13 & 0.13 & 0.12 \\
0.6 & 0.18 & 0.16 & 0.18 & 0.09 & 0.17 & 0.19 & -0.02 & 0.19 \\
0.7 & 0.27 & 0.23 & 0.24 & 0.14 & 0.30 & 0.29 & 0.00 & 0.11 \\
0.8 & 0.34 & 0.33 & 0.32 & 0.18 & 0.00 & 0.38 & -0.01 & 0.29 \\
0.9 & 0.15 & 0.43 & 0.45 & 0.24 & 0.51 & 0.54 & -0.02 & 0.40 \\
1.0 & 0.69 & -0.01 & 0.61 & 0.37 & 0.66 & 0.71 & 0.70 & 0.50 \\
\bottomrule
\end{tabular}
\label{tab:consistency_numerical}
\end{table}

\newpage

\section{Additional results}\label{app:additional_results}
\subsection{Ising model entropy estimation}\label{subsec:app_ising} 

The Ising model is a system consisting of particles arranged in a lattice. In our experiments, we consider a $L\times L$ square. A particle $i$ of the lattice is associated with a discrete value $\sigma_{i}\in\{-1,+1\}$ called spin and each pair of particles $ij$ is characterized by an interaction strength $J_{ij}$. With no external fields, these quantities determine the energy $E(\sigma)$ of the configuration $\sigma$: 

\begin{equation*}
E(\sigma)=\sum_{i,j}J_{ij}\sigma_i\sigma_j
\end{equation*}

In turns, the energy of a configuration determines its likelihood. In particular, the configurations of the Ising model follow a probability distribution $\fp_0$ parametrised by the temperature $T$, the Boltzmann constant $\boltzmann$ and the interaction strengths:

\begin{equation*}
    \fp_0(\sigma) = \tfrac{e^{-\beta E(\sigma)}}{Z(T)}
\end{equation*}

Where $Z(T)=\sum_{i}e^{-\beta E(\sigma_i)}$ and $\beta=(\boltzmann T)^{-1}$. In order to generate our dataset from $\fp_0$, we follow the Metropolis algorithm \citep{bhanot1988metropolis}:

\begin{algorithm}[H]
\caption{Metropolis Algorithm for 2D Ising Spin Glass}
\begin{algorithmic}[1]\label{alg:metropolis}
\STATE \textbf{Input:} Lattice size \( N \), interaction strengths \( J_{ij} \), temperature \( T \), number of iterations \( \text{iter\_max} \)
\STATE \textbf{Initialize:} Spin lattice \( \sigma \) with \( \sigma_{i,j} \in \{-1, +1\} \) randomly assigned
\FOR{iteration = 1 to \( \text{iter\_max} \)}
    \STATE Randomly select a lattice site \(i \)
    \STATE Compute the change in energy \( \Delta E \) if \( \sigma_{i,j} \) is flipped:
    \[
    \Delta E = 2 \, \sigma_i \sum_{j} J_{i,j} \, \sigma_{j}
    \]
    \STATE Generate a random number \( r \) uniformly distributed in \( [0, 1] \)
    \IF{ \( r < \exp\left(-\beta\Delta E\right) \) }
        \STATE Flip the spin: \( \sigma_i \leftarrow -\sigma_i \)
    \ENDIF
\ENDFOR
\STATE \textbf{Output:} Final spin configuration \( \sigma \)
\end{algorithmic}
\end{algorithm}

We compute the entropy of $\fp_0$ analytically, starting from the free energy $F$ per site of the lattice:

\begin{equation*}
    F(T) = -\boltzmann T\log\pf_T
\end{equation*}
Where $\pf_T$ is the partition function, which depends on the interaction horizontal and vertical interaction strength. For simplicity, we consider the same interaction strength $J=1$ for all neighboring particles, while we set it to zero for non neighboring particles. Under these assumptions, we can calculate $\log\lambda$ with a double integral\citep{onsager1944crystal}:

{\footnotesize\begin{equation*}
    \log\pf_T = \log 2 + \tfrac{1}{2\pi^2}\int_0^\pi\int_0^\pi\log(\cosh(2\beta J)\cosh(2\beta J)-\sinh(2\beta J)\cos(\theta_1)-\sinh(2\beta J)\cos(\theta_2))d\theta_1d\theta_2
\end{equation*}}

For simplicity, we also set $\boltzmann=1$. From $F$, we can calculate the entropy $H$ using the thermodynamic relation $H=-\tfrac{\partial F}{\partial T}$. We compute the integral numerically using the \textit{SciPy} Python package \citep{2020SciPy-NMeth} and we approximate $H$ as $H\approx \tfrac{F(T+\Delta T)-F(T-\Delta T)}{2\Delta T}$, with $\Delta T = 10^{-4}$.

For what concerns the \textsc{info-sedd} architecture, we keep a single model configuration for all temperatures, both for the model, which contains around $90k$ parameters, and for the diffusion. To get the entropy per site, we divide the estimates of the model by the number of particles in the configurations (400).

In this section, we numerically validate the performance of \infosedd\ by estimating entropy according to \Cref{alg:compute_entropy}.

\paragraph{Spin glasses experiments.} Entropy computation in Ising models enables insights on the thermodynamics properties of the system \citep{cincio2007entropy}, which can be used for scientific discovery in the domain where the Ising model is applied \citep{macy2024ising, schneidman2006weak, sherrington1975solvable}. 

\paragraph{Experimental setup} We consider a simplified Ising model applied to spin glasses \citep{sherrington1975solvable}. We do not include an external field, we set a unitary interaction strength for all the sites interactions, unitary Boltzmann's constant and a $20\times20$ square lattice. The entropy per site of this configuration can be computed in closed form \cite{onsager1944crystal}. We test our model by estimating the entropy per particle at linearly spaced temperatures from $1.0K$ to $4.0K$. We generate our dataset using the Metropolis-Hastings algorithm, with $10000$ samples for each temperature (see \Cref{alg:metropolis}). We post-process the output of \infosedd\ by dividing the entropy estimate by $400$ to report the entropy per site.

\begin{figure}
    \centering
    \includegraphics[width=0.75\linewidth]{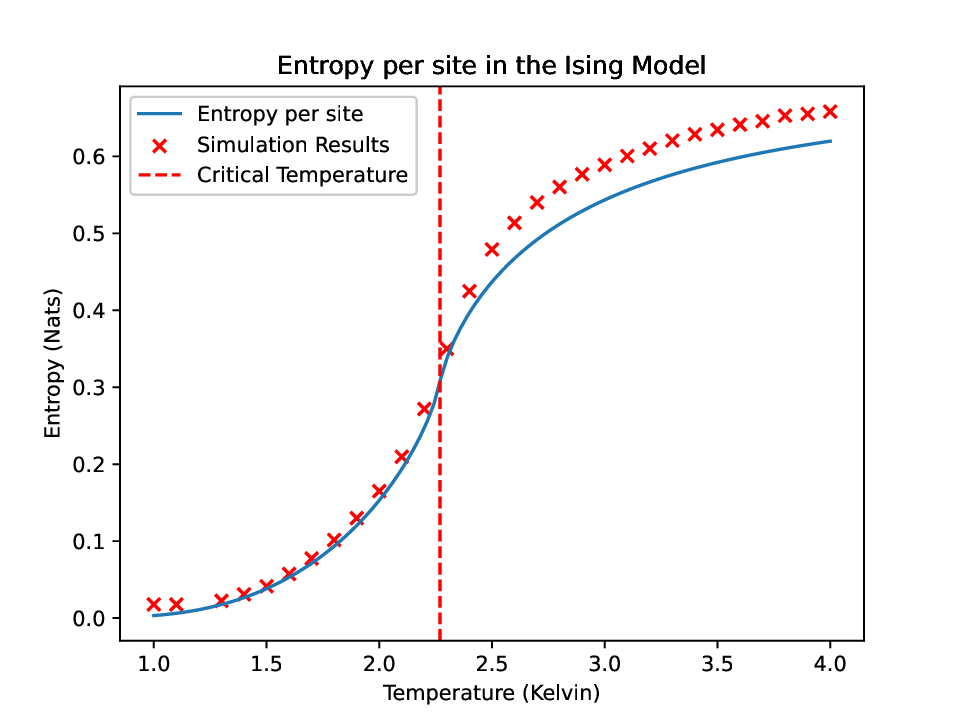}
    \caption{Ising model entropy estimates}
    \label{fig:ising}
\end{figure}

\paragraph{Results and Analysis} Variational estimators cannot estimate large \gls{KL} divergences reliably with limited samples sizes \citep{mcallester2020formal,song2019understanding}. In this scenario, instead, \infosedd\ accurately estimates large \gls{KL} divergences (\Cref{fig:ising}), performing particularly well at low temperatures where we need to estimate large \gls{KL} divergences.

\subsection{Comparison with plug-in estimators}

In this section, we discuss the limitations of plug-in estimators for discrete random variables. 
First, we analyze the approach by \citet{goebel}, where the authors use a Taylor expansion to estimate \gls{MI}. 
However, by construction, this estimator assumes a low \gls{MI} regime. 
This assumption makes the method inherently unsuitable for high-\gls{MI} estimation tasks, where higher-order terms in the expansion become non-negligible and induce severe bias.

We further analyze the approach of \citet{gao2017estimating}, who propose a variation of the KSG estimator to handle mixtures of continuous and discrete data. 
We benchmark this estimator on the high-dimensional synthetic benchmark, as in \Cref{tab:mi_comparison_std}, and on the extreme high-dimensional setting introduced in \Cref{sec:extremedim}. 
Results are reported in \Cref{tab:plugin_high_mi,tab:plugin_extreme_dim}.

\begin{table}[h]
\centering
\caption{Plug-in estimator (\citet{gao2017estimating}) on high-MI synthetic benchmarks with $D=\mathrm{MI}$.}
\label{tab:plugin_high_mi}
\begin{tabular}{l c}
\toprule
Setting & Estimated MI \\
\midrule
MI=10, D=10 & 11.56 \\
MI=20, D=20 & 9.33 \\
MI=30, D=30 & 10.55 \\
MI=40, D=40 & 2.87 \\
MI=50, D=50 & 4.18 \\
\bottomrule
\end{tabular}
\end{table}

We observe that while the estimator remains reasonably accurate for moderate MI (e.g., MI=10), its performance quickly deteriorates as dimensionality and dependence strength increase. 

We additionally evaluate the estimator in the extreme dimensionality setting with low ground-truth MI ($0.5$), as introduced in \Cref{sec:extremedim}.

\begin{table}[h]
\centering
\caption{Plug-in estimator (\citet{gao2017estimating}) in extreme dimensionality regimes with ground-truth MI $=0.5$.}
\label{tab:plugin_extreme_dim}
\begin{tabular}{l c c}
\toprule
\gls{MI} & D & Estimated \gls{MI} \\
\midrule
0.5 & 256  & 14.17 \\
0.5 & 1024 & 14.30 \\
\bottomrule
\end{tabular}
\end{table}

In this regime, the estimator overestimates the mutual information. This behavior highlights the instability of plug-in approaches in high-dimensional discrete settings.

Overall, these experiments confirm that plug-in estimators based on Taylor approximations or nearest-neighbor corrections are not robust to either high-MI or high-dimensional regimes. 
In contrast, \infosedd\ remains stable and accurate across both challenging scenarios.

\newpage
\section{Theoretical properties of \infosedd}\label{app:theory}

We highlight the main assumptions and setting in \Cref{sec:assumptions}, then in \Cref{app:vanishing_term} we proceed to construct an error bound on the \gls{KL} estimate of \infosedd\ under the established setting. 


\subsection{Assumptions and setting}\label{sec:assumptions}
We highlight the assumptions on the probability distributions and score networks.
We first define the class of functions $\mathcal{F}$ which restricts the class of score functions and score networks. 
In this work, we consider \gls{CTMC} with only sparse transition rate matrices $\fratesp$, which constrain the \gls{CTMC} to modify only one subcomponent at a time. This means that, when considering the \gls{KL} estimate as in \Cref{eq:infosedd}, only ratios of probability distributions $\tfrac{\fp_t(a)}{\fp_t(b)}$ with $a$ and $b$ within Hamming distance~1 are considered. The assumptions that we are going to make in this section reflect this characteristic. We define the set of pairs of samples at Hamming distance~1 as follows:
\[
\mathcal{S}:=\{(x',x'') \in \chi^D\times\chi^D \mid \hamming(x',x'')=1\}.
\]
Where $\chi^D$ is the support of $D$-dimensional random vectors
We impose two finite constants
\[
0<C_1<C_2<+\infty,
\]
and define
\[
\mathcal{F}:=\{f:\mathcal{S}\times[0,T]\to\mathbb{R} \mid\ C_1\leq ||f||_\infty\leq C_2\},
\]
where $||f||_\infty$ denotes the supremum norm over $\mathcal{S}\times[0,T]$. 

Throughout this appendix, we use a family of neural networks $T_\Theta\subset \mathcal{F}$ that are standard multilayer feedforward networks with bounded, nonconstant activation functions \citep{hornik1989multilayer}. We also fix a perturbation kernel $\fp_{t|0}$ and denote the score function for arbitrary probability distribution $p$ as 
\[
s^p(x_t)_{\hat{x}} = \frac{\fp_t(\hat{x})}{\fp_t(x_t)}, 
\]
where $\fp_t(x_t)=\sum_{x_0}\fp_{t|0}(x_t|x_0)p(x_0)$. 

We now state the assumptions on score networks and on score functions:
\begin{itemize}
    \item \textbf{A.1}: For all $\scorep\in T_\Theta$, $C_1\leq ||\scorep||_\infty\leq C_2$.
    \item \textbf{A.2}: For all $p$, the score function satisfies $C_1\leq ||s^p||_\infty\leq C_2$.
\end{itemize}
The lower bound $C_1>0$ avoids degeneracies, and ensures well-defined log terms in the loss.
We define the class of distribution $\mathcal{P}$ for which \textbf{A.2} is true as
\[
\mathcal{P} := \{p\, |C_1\leq ||s^p||_\infty\leq C_2\}
\]

{\textbf{A.1} and \textbf{A.2} are similar to Assumption 4.4 of \citet{ren2024discrete}, who assume that both the score function and the score networks are bounded. Similarly, \citet{chen2025convergence} set bound for the score network (Assumption 4.2) and for the true score function (Assumption 4.3).



\subsection{Error analysis of \infosedd}\label{app:vanishing_term}
 Let $p,q \in \mathcal{P}$ be two probability distributions satisfying the assumptions highlighted in \Cref{sec:assumptions}. Let $\theta,\phi\in\Theta$ be two { arbitrary} sets of parameters for the score networks $\scorep$, $\scoreq$. 

 \[\perr = || \scorep - s^p \bigr||_\infty , \qquad \qerr = || \scoreq - s^q ||_\infty. \]
 
  Then, let us define the functional $\isedd$, parameterised with arbitrary functions $f,g \in \mathcal{F}$ and perturbation kernel $\fp_{t|0}$ as

\begin{equation*}
    \isedd(f,g;x_0) = \int_0^T \expected_{x_t\sim \fp_{t|0}(\cdot|x_0)} \sum_{\hat{x}\neq x}\fratesp(x_t, \hat{x})({K(f(x_t)_{\hat{x}}})+{f(x_t)_{\hat{x}}}-{g(x_t)_{\hat{x}}}\log{g(x_t)_{\hat{x}}})dt
\end{equation*}

Then, the \infosedd\ estimator for the \gls{KL} is 
\begin{equation*}
    \expected_{x\sim p}\isedd(\scorep,\scoreq;x)
\end{equation*}

{ The true \gls{KL} divergence, instead is exactly

\begin{equation*}
    \KL{p}{q} = \KL{\fp_T}{\fq_T}+\expected_{x\sim p}\isedd\left(s^p,s^q;x\right)
\end{equation*}
}

Then, we chracterise the error of \infosedd\ as

{\scriptsize\begin{equation}\label{eq:total_error}
    \left|\expected_{x\sim p}\isedd(\scorep,\scoreq;x)-\KL{\fp_T}{\fq_T}-\expected_{x\sim p}\isedd\left(s^p,s^q;x\right)\right| \leq \left|\expected_{x\sim p}\isedd(\scorep,\scoreq;x)-\expected_{x\sim p}\isedd\left(s^p,s^q;x\right)\right| + \KL{\fp_T}{\fq_T}
\end{equation}}

We denote $\absorb^D$ the element in $\chi^D$ such that $\absorb^D_i = \absorb\, \forall\, i$ and we now focus first on the second term of the right hand side of \Cref{eq:total_error}
\begin{equation*}
    \KL{\fp_T}{\fq_T} = \sum_x\fp_T(x)\log\tfrac{\fp_T(x)}{\fq_T(x)}
\end{equation*}


{ Firstly, note that for any $t\in(0,T]$, $\fp_t(\absorb^D)=\fq_t(\absorb^D)$, and as a consequence $\log\tfrac{\fp_t(\absorb^D)}{\fq_t(\absorb^D)} = \log (1) = 0$. This means that we can write}

{
\begin{align*}
    \KL{\fp_T}{\fq_T} &= \sum_x\fp_T(x)\log\tfrac{\fp_T(x)}{\fq_T(x)} \\
    &= \sum_{x\neq\absorb^D}\fp_T(x)\log\tfrac{\fp_T(x)}{\fq_T(x)} \\
    &\leq \sum_{x\neq\absorb^D}\fp_T(x)\log\tfrac{1}{\fq_T(x)} \\
    &= \sum_{x\neq\absorb^D}\fp_T(x)\log\sum_{\hat{x}}\tfrac{\fq_T(\hat{x})}{\fq_T(x)}.
\end{align*}
Note that we cannot bound $\tfrac{\fq_T(\hat{x})}{\fq_T(x)}$ by simply applying \textbf{A.2}, since the sum involves terms such that $\hamming(x,\hat x)>1$. We can however construct an ordered sequence of points $S(x,\hat{x})$ where $x$ is the first element, $\hat{x}$ is the last element and $\forall\,i,j$ such that $0\leq i<j<\hamming(x,\hat{x})$ it holds $\hamming(S(x,\hat{x})_j,S(x,\hat{x})_i)=j-i$. This allows us to write

\begin{align*}
 \tfrac{\fq_T(\hat{x})}{\fq_T(x)}& =\prod_{i=0}^{\hamming(x,\hat{x})-2}\tfrac{\fq_T(S(x,\hat{x})_{i+1})}{\fq_T(S(x,\hat{x})_i)} \\
 &=\prod_{i=0}^{\hamming(x,\hat{x})-2}s^q(S(x,\hat{x})_i)_{S(x,\hat{x})_{i+1}}\\
 &\leq\prod_{i=0}^{\hamming(x,\hat{x})-2} C_2 \\
 &\leq C_2^{D}
\end{align*}
}

We can now go back to $\KL{\fp_T}{\fq_T}$,

\begin{align*}
    \KL{\fp_T}{\fq_T} &\leq \sum_{x\neq\absorb^D}\fp_T(x)\log\sum_{\hat{x}}\tfrac{\fq_T(\hat{x})}{\fq_T(x)} \\
    &\leq \sum_{x\neq\absorb^D}\fp_T(x)\log\sum_{\hat{x}} C_2^D \\
    &= \sum_{x\neq\absorb^D}\fp_T(x)\log |\chi|^DC_2^D \\
    &= (1-\fp_T(\emptyset^D)) (D\log |\chi|C_2).
\end{align*}

We can now focus on the { first term of the right hand side of \Cref{eq:total_error}
\[
\left|\expected_{x\sim p}\isedd(\scorep,\scoreq;x)-\expected_{x\sim p}\isedd\left(s^p,s^q;x\right)\right|
\]}

{ First, we decompose each term and use triangle inequality} 
\begin{align}
    &\left|\expected_{x\sim p}\isedd(\scorep,\scoreq;x)
      - \expected_{x\sim p}\isedd\!\left(s^p,
        s^q;x\right)\right| \nonumber \\
    &\quad= \left|
      \expected_{x\sim p}\Biggl(\isedd(\scorep,\scoreq;x)
        - \isedd\!\left(s^p,
          s^q;x\right)\Biggr)
      \right|  \nonumber \\
    &\quad= \Biggl|\expected_{x\sim p}\int_0^T 
      \expected_{x_t\sim \fp_{t|0}(\cdot|x_0)} 
      \sum_{\hat{x}\neq x}\fratesp(x_t, \hat{x})
      \Biggl(
        K\!\bigl(\scorep(x_t)_{\hat{x}}\bigr)
        - K\!\left(s^p(x_t)_{\hat{x}}\right) \nonumber \\
    &\qquad\qquad\qquad\quad
        + \scoreq(x_t)_{\hat{x}}
        - s^q(x_t)_{\hat{x}}
        - \scorep(x_t)_{\hat{x}}
          \log\!\bigl(\scoreq(x_t)_{\hat{x}}\bigr) \nonumber \\
    &\qquad\qquad\qquad\quad
        + s^p(x_t)_{\hat{x}}
          \log\!\left(s^q(x_t)_{\hat{x}}\right)
      \Biggr)\,dt \Biggr| \nonumber  \\
    &\quad\leq \expected_{x\sim p} \int_0^T 
      \expected_{x_t\sim \fp_{t|0}(\cdot|x_0)} 
      \sum_{\hat{x}\neq x}\fratesp(x_t, \hat{x})
      \Biggl(
        \Biggl|
          K\!\bigl(\scorep(x_t)_{\hat{x}}\bigr)
          - K\!\left(s^p(x_t)_{\hat{x}}\right)
        \Biggr| \label{eq:error_terms} \\
    &\qquad\qquad\qquad\quad
        + \Biggl|
          \scoreq(x_t)_{\hat{x}}
          - s^q(x_t)_{\hat{x}}
        \Biggr| \nonumber \\
    &\qquad\qquad\qquad\quad
        + \Biggl|
          \scorep(x_t)_{\hat{x}}
          \log\!\bigl(\scoreq(x_t)_{\hat{x}}\bigr)
          - s^p(x_t)_{\hat{x}}
            \log\!\left(s^q(x_t)_{\hat{x}}\right)
        \Biggr|
      \Biggr)\,dt \nonumber
\end{align}

We use the inequality $\log(a) \leq a-1$ to proceed with the first term of \Cref{eq:error_terms},
\begin{align*}
     \left|K\!\bigl(\scorep(x_t)_{\hat{x}}\bigr)
        - K\!\left(s^p(x_t)_{\hat{x}}\right)\right| &= \left|\scorep(x_t)_{\hat{x}}(\log(\scorep(x_t)_{\hat{x}}) - 1)-\left(s^p(x_t)_{\hat{x}}\right)\left(\log\left(s^p(x_t)_{\hat{x}}\right)-1\right)\right| \\
        &\leq \left|\scorep(x_t)_{\hat{x}}\log(\scorep(x_t)_{\hat{x}})-\left(s^p(x_t)_{\hat{x}}\right)\log\left(s^p(x_t)_{\hat{x}}\right)\right| + \perr \\
        &\leq \max(s^p(x_t)_{\hat{x}},\scorep(x_t)_{\hat{x}})\left| \log(\scorep(x_t)_{\hat{x}})-\log\left(s^p(x_t)_{\hat{x}}\right)\right| +\perr \\
        &\leq C_2\left|\log\left(\tfrac{\scorep(x_t)_{\hat{x}}}{s^p(x_t)_{\hat{x}}}\right)\right| + \perr\\
        &\leq C_2\left|\left(\tfrac{\max(\scorep(x_t)_{\hat{x}},s^p(x_t)_{\hat{x}})}{\min(\scorep(x_t)_{\hat{x}},s^p(x_t)_{\hat{x}})}-1\right)\right| +\perr \\
        &\leq C_2\left|\left(\tfrac{\max(\scorep(x_t)_{\hat{x}},s^p(x_t)_{\hat{x}})-\min(\scorep(x_t)_{\hat{x}},s^p(x_t)_{\hat{x}})}{\min(\scorep(x_t)_{\hat{x}},s^p(x_t)_{\hat{x}})}\right)\right| +\perr \\
        &\leq \tfrac{C_2}{\min(\scorep(x_t)_{\hat{x}},s^p(x_t)_{\hat{x}})}\left|\left(\scorep(x_t)_{\hat{x}}-s^p(x_t)_{\hat{x}}\right)\right| + \perr \\
        &\leq \tfrac{C_2}{C_1}\left|\left(\scorep(x_t)_{\hat{x}}-s^p(x_t)_{\hat{x}}\right)\right| + \perr \\
        &\leq \tfrac{C_2}{C_1}\perr + \perr \\
        &= \perr(1+\tfrac{C_2}{C_1}) \\
\end{align*}

Then, we continue with the second term of \Cref{eq:error_terms}, which is simply 
\begin{equation*}
    \left|\scoreq(x_t)_{\hat{x}}
        - s^q(x_t)_{\hat{x}}\right| \leq \qerr
\end{equation*}

Finally, we write the term of \Cref{eq:error_terms}
\begin{align*}
  &\Bigl|\scorep(x_t)_{\hat{x}}\log\!\bigl(\scoreq(x_t)_{\hat{x}}\bigr)
    - s^p(x_t)_{\hat{x}}
      \log\!\biggl(s^q(x_t)_{\hat{x}} \biggr)\Bigr| \\
  &\qquad\leq
    \max\!\Bigl(\scorep(x_t)_{\hat{x}},s^p(x_t)_{\hat{x}}\Bigr)
    \Bigl|\log s^q(x_t)_{\hat{x}}
      -\log\bigl(\scoreq(x_t)_{\hat{x}}\bigr)\Bigr| \\
  &\qquad\leq
    C_2 \Bigl|\log s^q(x_t)_{\hat{x}}
      -\log\bigl(\scoreq(x_t)_{\hat{x}}\bigr)\Bigr| \\
  &\qquad\leq
    \tfrac{C_2}{C_1} \Bigl|s^q(x_t)_{\hat{x}}
      -\scoreq(x_t)_{\hat{x}}\Bigr| \\
  &\qquad\leq
    \tfrac{C_2}{C_1}\,\qerr
\end{align*}

Combining the results we obtain
\begin{align*}
    &\left|\expected_{x\sim p}\isedd(\scorep,\scoreq;x)-\expected_{x\sim p}\isedd\left(s^p,s^q;x\right)\right| \\ &\leq \Biggl|\int_0^T \expected_{x_t\sim \fp_{t|0}(\cdot|x_0)} 
      \sum_{\hat{x}\neq x_t}\fratesp(x_t, \hat{x})\left((\perr+\qerr)\left(1+\tfrac{C_2}{C_1}\right)\right)dt \\
      & \leq \bar{\sigma}(T)D|\chi|\left((\perr+\qerr)\left(1+\tfrac{C_2}{C_1}\right)\right)
\end{align*}

where the factor $D|\chi|$ is the number of non-zero entries in $\fratesp$ using the absorb configuration. 
We can finally put together everything to get the error bound for \Cref{eq:total_error} 

\begin{align*}
& \left|\expected_{x\sim p}\isedd(\scorep,\scoreq;x)-\KL{\fp_T}{\fq_T}-\expected_{x\sim p}\isedd\left(s^p,s^q;x\right)\right| \\ & \leq \left|\expected_{x\sim p}\isedd(\scorep,\scoreq;x)-\expected_{x\sim p}\isedd\left(s^p,s^q;x\right)\right| + \KL{\fp_T}{\fq_T} 
\\ &\leq 
    \bar{\sigma}(T)D|\chi|\left((\perr+\qerr)\left(1+\tfrac{C_2}{C_1}\right)\right)+(1-\fp_T(\emptyset^D)) (D\log |\chi|C_2) 
\end{align*}

Note that $\fp_T(\absorb^D)$ approaches 1 exponentially in $T$ (Equation 14 of \citet{lou2024discrete}). The other term, instead, shows that the error on the \gls{KL} increases linearly with the error of the score network for fixed $T$ and does not depend on terms exponential on the \gls{KL} divergence. 

%
\subsection{Discussion on consistency}
As presented in Appendix E, the error of \infosedd\ decomposes in two terms: one proportional to the error of the score networks, and one proportional to $\fp_T(\absorb^D)$. 
\begin{align}\label{eq:error_infosedd}
& \left|\expected_{x\sim p}\isedd(\scorep,\scoreq;x)-\KL{\fp_T}{\fq_T}-\expected_{x\sim p}\isedd\left(s^p,s^q;x\right)\right| \nonumber \\ & \leq \left|\expected_{x\sim p}\isedd(\scorep,\scoreq;x)-\expected_{x\sim p}\isedd\left(s^p,s^q;x\right)\right| + \KL{\fp_T}{\fq_T} \nonumber
\\ &\leq 
    \bar{\sigma}(T)D|\chi|\left((\perr+\qerr)\left(1+\tfrac{C_2}{C_1}\right)\right)+(1-\fp_T(\emptyset^D)) (D\log |\chi|C_2) 
\end{align}
 The second term of \Cref{eq:error_infosedd} cannot be reduced with better score networks, as it only depends on the \gls{CTMC} and the data distribution. However, it decreases exponentially in $\bar{\sigma}(T)$ (Equation 14 of \citet{lou2024discrete}), establishing a trade-off with the first term, which only increases linearly with $\bar{\sigma}(T)$.
 For this reason, we focus on the first term. Note that our final KL estimator \Cref{eq:infosedd} is a continuous functional of the score networks $\scorep,\scoreq$. By theorem 3.4 of \citet{lou2024discrete}, the DWDSE loss is an equivalent objective compared to the score entropy loss. This allows us to apply proposition 3.2 of \citet{lou2024discrete}, for which, given enough samples and model capacity, the score network $s_\theta^p$ and $s_\phi^q$ converge to their true values $s^p$ and $s^q$.
Our final KL estimator \Cref{eq:infosedd} is a continuous functional of these score networks. By the Continuous Mapping Theorem (theorem 1.9.5 of \citet{vaart1997weak}), assuming as in \citet{lou2024discrete} that $s_\theta^p$ and $s_\phi^q$ converge in probability to their true values $s^p$ and $s^q$, the output of the functional (our KL estimate) must also converge in probability to the true KL divergence minus the truncation bias.
\subsection{Discussion on bias}
Consider the \infosedd\ estimator

\[
\alpha = \expected\left[\int_0^T\sum_{x\neq \fproc_{t}} \fratesp_{t}(\fproc_{t},x)\left(K\left(s^p(\fproc_{t})_x\right)+s^q(\fproc_{t})_x-s^p(\fproc_{t})_x\log s^q(\fproc_{t})_x\right) dt\right]
\]

and its empirical version over a collection of samples $x_0^i\sim \pdata$

\[
\hat{\alpha} = \frac{1}{N}\sum_{i=1}^{N}\left[\int_0^T\sum_{x\neq x_t^i} \fratesp_{t}(x_t^i,x)\left(K\left(s^p(x_t^i)_x\right)+s^q(x_t^i)_x-s^p(x_t^i)_x\log s^q(x_t^i)_x\right) dt\right].
\]

Let us denote $\phi(x_t)=\sum_{x\neq x_t^i} \fratesp_{t}(x_t^i,x)\left(K\left(s^p(x_t^i)_x\right)+s^q(x_t^i)_x-s^p(x_t^i)_x\log s^q(x_t^i)_x\right)$, then we can write its expectation as
\begin{align*}
    \expected[\hat{\alpha}] &= \expected\left[\frac{1}{N}\sum_{i=1}^N\int_0^T\phi(x^i_t)dt\right] \\
    &=\frac{1}{N}\sum_{i=1}^N\expected\left[\int_0^T\phi(x^i_t) dt\right]
\end{align*}

Note that $\expected\left[\int_0^T\phi(x^i_t) dt\right]$ is the true $\alpha$. As a consequence $\expected[\hat{\alpha}]=\alpha$, which means that \infosedd\ is an unbiased estimator for $\alpha$. Note, however, that we still have the truncation bias which is bounded by
\[
|\KL{\pdata}{\qdata} -\alpha| \leq (1-\fp_T(\emptyset^D)) (D\log |\chi|C_2).
\]
This source of bias is object of tradeoff with the other term
\[
\bar{\sigma}(T)D|\chi|\left((\perr+\qerr)\left(1+\tfrac{C_2}{C_1}\right)\right)
\]
since increasing $T$ would reduce the former but expand the latter. However, the term $(1-\fp_T(\emptyset^D))$ decreases exponentially in $\bar{\sigma}(T)$ (Equation 14 of \citet{lou2024discrete}). This means that the truncation bias can be drastically reduced without affecting the other error term which only has a linear dependence on $\bar{\sigma}(T)$. 

%% file: tables/memory_table.tex
\begin{table}[ht]
\centering
\scriptsize
\begin{tabular}{lllllllll}
\toprule
Estimator & \infosedd & \textsc{gan-dime} & \textsc{hd-dime} & \textsc{kl-dime} & \textsc{minde} & \textsc{mine} & \textsc{nwj} & \textsc{smile} \\
\midrule
MI=10, D=10 & \textbf{346.37} & 645.81 & 645.81 & 645.81 & 440.51 & 645.81 & 645.81 & 645.81 \\
MI=20, D=20 & \textbf{687.22} & 1301.98 & 1301.98 & 1301.98 & 876.44 & 1301.98 & 1301.98 & 1301.98 \\
MI=30, D=30 & \textbf{1005.66} & 1913.15 & 1913.15 & 1913.15 & 1286.35 & 1913.15 & 1913.15 & 1913.15 \\
MI=40, D=40 & \textbf{1345.94} & 2564.33 & 2564.33 & 2564.33 & 1726.25 & 2564.32 & 2564.32 & 2564.32 \\
MI=50, D=50 & \textbf{1666.30} & 3180.50 & 3180.50 & 3180.50 & 2138.41 & 3180.49 & 3180.49 & 3180.49 \\
\bottomrule
\end{tabular}
\caption{Average maximum memory usage over one epoch for different estimators over different \gls{MI} estimation tasks. Results are in MB. Lowest memory usages are written in bold.}
\label{tab:memory}
\end{table}

%% file: tables/runtime_table.tex
\begin{table}[ht]
\centering
\scriptsize
\begin{tabular}{lllllllll}
\toprule
Estimator & \infosedd & \textsc{gan-dime} & \textsc{hd-dime} & \textsc{kl-dime} & \textsc{minde} & \textsc{mine} & \textsc{nwj} & \textsc{smile} \\
\midrule
MI=10, D=10 & 2.40 & 2.89 & 2.90 & 2.86 & \textbf{2.16} & 2.89 & 2.96 & 2.94 \\
MI=20, D=20 & 3.34 & 3.99 & 4.02 & 3.98 & \textbf{2.62} & 4.03 & 4.03 & 4.00 \\
MI=30, D=30 & 4.51 & 5.56 & 5.54 & 5.57 & \textbf{3.37} & 5.57 & 5.57 & 5.61 \\
MI=40, D=40 & 5.96 & 7.34 & 7.33 & 7.36 & \textbf{4.00} & 7.39 & 7.42 & 7.37 \\
MI=50, D=50 & 7.06 & 8.71 & 8.69 & 8.71 & \textbf{4.68} & 8.76 & 8.73 & 8.72 \\
\bottomrule
\end{tabular}
\caption{Average runtime for one epoch for different estimators over different \gls{MI} estimation tasks. Results are in seconds. Best runtimes are written in bold.}
\label{tab:runtime}
\end{table}

%% file: tables/synthetic/low_dim.tex
\begin{center}
{\scriptsize \begin{tabular}{lrrrrrrrr}
\toprule
Estimator & \infosedd & \textsc{gan-dime} & \textsc{hd-dime} & \textsc{kl-dime} & \textsc{minde} & \textsc{mine} & \textsc{nwj} & \textsc{smile} \\
\midrule
MI=0, D=10 & 0.116 & 0.026 & 0.019 & 0.045 & 2.899 & \textbf{0.018} & 0.022 & 0.031 \\
MI=1, D=10 & 0.918 & 1.088 & 0.868 & 0.975 & 2.495 & \textbf{0.999} & 1.015 & 1.019 \\
MI=2, D=10 & 1.764 & 2.073 & 1.801 & 1.948 & 2.389 & \textbf{2.009} & 1.995 & 2.361 \\
MI=3, D=10 & 2.657 & 2.820 & 3.098 & 3.057 & 3.047 & \textbf{2.960} & 2.935 & 3.555 \\
MI=4, D=10 & 4.121 & \textbf{4.013} & 3.517 & 4.094 & 3.600 & 4.049 & 3.908 & 4.697 \\
MI=5, D=10 & \textbf{4.953} & 4.905 & 4.770 & 4.652 & 4.432 & 4.801 & 4.692 & 2.593 \\
\bottomrule
\end{tabular}}
\captionof{table}{Results for the low dimensional synthetic benchmark for all estimators. Best estimator in bold.}
\label{tab:mi_comparison_low_dim}
\end{center}

%% file: tables/synthetic/big_supp.tex
{\small\begin{center}
\captionof{table}{Results of the ablation study on $|\chi|$, with $D$ fixed to 10 and \gls{MI}=0.5. Best estimates in bold.\\}
\begin{tabular}{lrrrrrrrr}
\toprule
Estimator & \infosedd & \textsc{gan-dime} & \textsc{hd-dime} & \textsc{kl-dime} & \textsc{minde} & \textsc{mine} & \textsc{nwj} & \textsc{smile} \\
$\chi$ &  &  &  &  &  &  &  &  \\
\midrule
12 & \textbf{0.478} & 1.457 & 1.211 & 1.527 & 0.254 & 1.096 & 1.181 & 1.597 \\
36 & \textbf{0.486} & 1.610 & 1.323 & 1.507 & 0.008 & 1.093 & 1.128 & 1.600 \\
132 & \textbf{0.498} & 1.490 & 1.188 & 1.511 & 0.005 & 1.170 & 1.245 & 1.818 \\
\bottomrule
\end{tabular}

\label{tab:big_supp}
\end{center}}

%% file: sections_iclr/bibliography.bib
@inproceedings{lou2024discrete,
author = {Lou, Aaron and Meng, Chenlin and Ermon, Stefano},
title = {Discrete diffusion modeling by estimating the ratios of the data distribution},
year = {2024},
publisher = {JMLR.org},
booktitle = {Proceedings of the 41st International Conference on Machine Learning},
articleno = {1333},
numpages = {30},
location = {Vienna, Austria},
series = {ICML'24}
}

@inproceedings{franzese2023minde,
  title={MINDE: Mutual information neural diffusion estimation},
  author={Franzese, Giulio and Bounoua, Mustapha and Michiardi, Pietro},
  booktitle={International Conference on Learning Representations},
  volume={2024},
  pages={16685--16716},
  year={2024}
}

@inproceedings{
kholkin2025infobridgemutualinformationestimation,
title={InfoBridge: Mutual Information estimation via Bridge Matching},
author={Sergei Kholkin and Ivan Butakov and Evgeny Burnaev and Nikita Gushchin and Alexander Korotin},
booktitle={The Fourteenth International Conference on Learning Representations},
year={2026},
url={https://openreview.net/forum?id=y8Kzu9SKpv}
}

@inproceedings{
sun2023scorebasedcontinuoustimediscretediffusion,
title={Score-based Continuous-time Discrete Diffusion Models},
author={Haoran Sun and Lijun Yu and Bo Dai and Dale Schuurmans and Hanjun Dai},
booktitle={The Eleventh International Conference on Learning Representations },
year={2023},
url={https://openreview.net/forum?id=BYWWwSY2G5s}
}

@book{hanson2007applied,
  title={Applied stochastic processes and control for jump-diffusions: modeling, analysis and computation},
  author={Hanson, Floyd B},
  year={2007},
  publisher={SIAM}
}

@article{gao2017estimating,
  title={Estimating mutual information for discrete-continuous mixtures},
  author={Gao, Weihao and Kannan, Sreeram and Oh, Sewoong and Viswanath, Pramod},
  journal={Advances in neural information processing systems},
  volume={30},
  year={2017}
}

@article{lee2024benchmarksuiteevaluatingneural,
  title={A benchmark suite for evaluating neural mutual information estimators on unstructured datasets},
  author={Lee, Kyungeun and Rhee, Wonjong},
  journal={Advances in Neural Information Processing Systems},
  volume={37},
  pages={46319--46338},
  year={2024}
}

@inproceedings{holderrieth2024generator,
  title={Generator matching: Generative modeling with arbitrary markov processes},
  author={Holderrieth, Peter and Havasi, Marton and Yim, Jason and Shaul, Neta and Gat, Itai and Jaakkola, Tommi and Karrer, Brian and Chen, Ricky TQ and Lipman, Yaron},
  booktitle={International Conference on Learning Representations},
  volume={2025},
  pages={52153--52219},
  year={2025}
}

@article{fabbri2021summeval,
  title={Summeval: Re-evaluating summarization evaluation},
  author={Fabbri, Alexander R and Kry{\'s}ci{\'n}ski, Wojciech and McCann, Bryan and Xiong, Caiming and Socher, Richard and Radev, Dragomir},
  journal={Transactions of the Association for Computational Linguistics},
  volume={9},
  pages={391--409},
  year={2021},
  publisher={MIT Press One Rogers Street, Cambridge, MA 02142-1209, USA journals-info~…}
}

@inproceedings{
schiff2024caduceus,
title={Caduceus: Bi-Directional Equivariant Long-Range {DNA} Sequence Modeling},
author={Yair Schiff and Chia Hsiang Kao and Aaron Gokaslan and Tri Dao and Albert Gu and Volodymyr Kuleshov},
booktitle={Forty-first International Conference on Machine Learning},
year={2024},
url={https://openreview.net/forum?id=mk3A5IUdn8}
}

@article{bernard2010tc,
  title={TC-motifs at the TATA-box expected position in plant genes: a novel class of motifs involved in the transcription regulation},
  author={Bernard, Virginie and Brunaud, V{\'e}ronique and Lecharny, Alain},
  journal={BMC genomics},
  volume={11},
  pages={1--15},
  year={2010},
  publisher={Springer}
}

@inproceedings{ren2024discrete,
  title={How discrete and continuous diffusion meet: Comprehensive analysis of discrete diffusion models via a stochastic integral framework},
  author={Ren, Yinuo and Chen, Haoxuan and Rotskoff, Grant and Ying, Lexing},
  booktitle={International Conference on Learning Representations},
  volume={2025},
  pages={42904--42941},
  year={2025}
}

@inproceedings{song2019,
    author = {Song, Yang and Ermon, Stefano},
    booktitle = {Advances in Neural Information Processing Systems},
    editor = {H. Wallach and H. Larochelle and A. Beygelzimer and F. d' Alch\'{e}-Buc and E. Fox and R. Garnett},
    pages = {},
    publisher = {Curran Associates, Inc.},
    title = {Generative Modeling by Estimating Gradients of the Data Distribution},
    volume = {32},
    year = {2019}
}

@article{franzese2022,
  title={How much is enough? a study on diffusion times in score-based generative models},
  author={Franzese, Giulio and Rossi, Simone and Yang, Lixuan and Finamore, Alessandro and Rossi, Dario and Filippone, Maurizio and Michiardi, Pietro},
  journal={Entropy},
  year={2023},
  publisher={MDPI}
}

@inproceedings{kong2023information,
  title={Information-Theoretic Diffusion},
  author={Kong, Xianghao and Brekelmans, Rob and Ver Steeg, Greg},
  booktitle={International Conference on Learning Representations},
  year={2022}
}

@inproceedings{
hjelm2019learning,
title={Learning deep representations by mutual information estimation and maximization},
author={R Devon Hjelm and Alex Fedorov and Samuel Lavoie-Marchildon and Karan Grewal and Phil Bachman and Adam Trischler and Yoshua Bengio},
booktitle={International Conference on Learning Representations},
year={2019},
}

@InProceedings{belghazi2018mine,
  title = 	 {Mutual Information Neural Estimation},
  author =       {Belghazi, Mohamed Ishmael and Baratin, Aristide and Rajeshwar, Sai and Ozair, Sherjil and Bengio, Yoshua and Courville, Aaron and Hjelm, Devon},
  booktitle = 	 {Proceedings of the 35th International Conference on Machine Learning},
  year = 	 {2018},
}

@inproceedings{song2019understanding,
  title={Understanding the Limitations of Variational Mutual Information Estimators},
  author={Song, Jiaming and Ermon, Stefano},
  booktitle={International Conference on Learning Representations},
  year={2019}
}

@article{Karbowski_2024,
   title={Information Thermodynamics: From Physics to Neuroscience},
   volume={26},
   ISSN={1099-4300},
   url={http://dx.doi.org/10.3390/e26090779},
   DOI={10.3390/e26090779},
   number={9},
   journal={Entropy},
   publisher={MDPI AG},
   author={Karbowski, Jan},
   year={2024},
   month=sep, pages={779} }

@inproceedings{Eckford_2016,
   title={Finite-state channel models for signal transduction in neural systems},
   url={http://dx.doi.org/10.1109/ICASSP.2016.7472889},
   DOI={10.1109/icassp.2016.7472889},
   booktitle={2016 IEEE International Conference on Acoustics, Speech and Signal Processing (ICASSP)},
   publisher={IEEE},
   author={Eckford, Andrew W. and Loparo, Kenneth A. and Thomas, Peter J.},
   year={2016},
   month=mar, pages={6300–6304} }

@inproceedings{
brekelmans2023improving,
title={Improving Mutual Information Estimation with Annealed and Energy-Based Bounds},
author={Rob Brekelmans and Sicong Huang and Marzyeh Ghassemi and Greg Ver Steeg and Roger Baker Grosse and Alireza Makhzani},
booktitle={International Conference on Learning Representations},
year={2022},
}

@article{federici2023effectiveness,
  title={On the Effectiveness of Hybrid Mutual Information Estimation},
  author={Federici, Marco and Ruhe, David and Forr{\'e}, Patrick},
  journal={arXiv preprint arXiv:2306.00608},
  year={2023}
}

@article{letizia2024mutual,
  title={Mutual Information Estimation via $ f $-Divergence and Data Derangements},
  author={Letizia, Nunzio Alexandro and Novello, Nicola and Tonello, Andrea M},
  journal={Advances in Neural Information Processing Systems},
  volume={37},
  pages={105114--105150},
  year={2024}
}

@article{rhodes2020telescoping,
  title={Telescoping density-ratio estimation},
  author={Rhodes, Benjamin and Xu, Kai and Gutmann, Michael U},
  journal={Advances in neural information processing systems},
  year={2020}
}

@inproceedings{wunder2021reverse,
  title={A reverse Jensen inequality result with application to mutual information estimation},
  author={Wunder, Gerhard and Gro{\ss}, Benedikt and Fritschek, Rick and Schaefer, Rafael F},
  booktitle={2021 IEEE Information Theory Workshop (ITW)},
  year={2021},
}

@article{oord2019representation,
  title={Representation learning with contrastive predictive coding},
  author={Oord, Aaron van den and Li, Yazhe and Vinyals, Oriol},
  journal={Advances in neural information processing systems},
  year={2018}
}

@inproceedings{mcallester2020formal,
  title={Formal limitations on the measurement of mutual information},
  author={McAllester, David and Stratos, Karl},
  booktitle={International Conference on Artificial Intelligence and Statistics},
  year={2020},
}

@inproceedings{nguyen2007neurips,
 author = {Nguyen, XuanLong and Wainwright, Martin J and Jordan, Michael},
 booktitle = {Advances in Neural Information Processing Systems},
 title = {Estimating divergence functionals and the likelihood ratio by penalized convex risk minimization},
 year = {2007}
}

@inproceedings{nowozin2016neurips,
author = {Nowozin, Sebastian and Cseke, Botond and Tomioka, Ryota},
title = {F-GAN: Training Generative Neural Samplers Using Variational Divergence Minimization},
year = {2016},
isbn = {9781510838819},
publisher = {Curran Associates Inc.},
address = {Red Hook, NY, USA},
booktitle = {Proceedings of the 30th International Conference on Neural Information Processing Systems},
pages = {271–279},
numpages = {9},
location = {Barcelona, Spain},
series = {NIPS'16}
}

@inproceedings{poole2019variational,
  title={On variational bounds of mutual information},
  author={Poole, Ben and Ozair, Sherjil and Van Den Oord, Aaron and Alemi, Alex and Tucker, George},
  booktitle={International Conference on Machine Learning},
  year={2019}
}

@article{pizer1987adaptive,
  title={Adaptive histogram equalization and its variations},
  author={Pizer, Stephen M and Amburn, E Philip and Austin, John D and Cromartie, Robert and Geselowitz, Ari and Greer, Trey and ter Haar Romeny, Bart and Zimmerman, John B and Zuiderveld, Karel},
  journal={Computer vision, graphics, and image processing},
  volume={39},
  number={3},
  pages={355--368},
  year={1987},
  publisher={Elsevier}
}

@article{moon1995estimation,
  title={Estimation of mutual information using kernel density estimators},
  author={Moon, Young-Il and Rajagopalan, Balaji and Lall, Upmanu},
  journal={Physical Review E},
  volume={52},
  number={3},
  pages={2318},
  year={1995},
  publisher={APS}
}

@article{kraskov2004estimating,
  title={Estimating mutual information},
  author={Kraskov, Alexander and St{\"o}gbauer, Harald and Grassberger, Peter},
  journal={Physical review E},
  volume={69},
  number={6},
  pages={066138},
  year={2004},
  publisher={APS}
}

@inproceedings{alemi2019deep,
  title={Deep Variational Information Bottleneck},
  author={Alemi, Alexander A and Fischer, Ian and Dillon, Joshua V and Murphy, Kevin},
  booktitle={International Conference on Learning Representations},
  year={2016}
}

@article{chen2016infogan,
  title={Infogan: Interpretable representation learning by information maximizing generative adversarial nets},
  author={Chen, Xi and Duan, Yan and Houthooft, Rein and Schulman, John and Sutskever, Ilya and Abbeel, Pieter},
  journal={Advances in neural information processing systems},
  volume={29},
  year={2016}
}

@inproceedings{zhao2018information,
  title={A Lagrangian Perspective on Latent Variable Generative Models},
  author={Zhao, Shengjia and Song, Jiaming and Ermon, Stefano},
  booktitle={Proc. 34th Conference on Uncertainty in Artificial Intelligence},
  year={2018}
}

@article{alemi2019gilbo,
  title={GILBO: One metric to measure them all},
  author={Alemi, Alexander A and Fischer, Ian},
  journal={Advances in Neural Information Processing Systems},
  volume={31},
  year={2018}
}

@inproceedings{huang2020evaluating,
  title={Evaluating lossy compression rates of deep generative models},
  author={Huang, Sicong and Makhzani, Alireza and Cao, Yanshuai and Grosse, Roger},
  booktitle={International Conference on Machine Learning},
  year={2020},
  organization={PMLR}
}

@article{shannon1948mathematical,
  title={A mathematical theory of communication},
  author={Shannon, Claude Elwood},
  journal={The Bell system technical journal},
  volume={27},
  number={3},
  pages={379--423},
  year={1948},
  publisher={Nokia Bell Labs}
}

@book{mackay2003information,
  title={Information theory, inference and learning algorithms},
  author={MacKay, David JC},
  year={2003},
  publisher={Cambridge university press}
}

@inproceedings{gao2015efficient,
  title={Efficient estimation of mutual information for strongly dependent variables},
  author={Gao, Shuyang and Ver Steeg, Greg and Galstyan, Aram},
  booktitle={Artificial intelligence and statistics},
  pages={277--286},
  year={2015},
  organization={PMLR}
}

@article{barber2004algorithm,
  title={The im algorithm: a variational approach to information maximization},
  author={Barber, David and Agakov, Felix},
  journal={Advances in neural information processing systems},
  volume={16},
  number={320},
  pages={201},
  year={2004}
}

@article{papamakarios2017masked,
  title={Masked autoregressive flow for density estimation},
  author={Papamakarios, George and Pavlakou, Theo and Murray, Iain},
  journal={Advances in neural information processing systems},
  volume={30},
  year={2017}
}

@article{letizia2022copula,
  title={Copula density neural estimation},
  author={Letizia, Nunzio A and Novello, Nicola and Tonello, Andrea M},
  journal={IEEE Transactions on Neural Networks and Learning Systems},
  year={2025},
  publisher={IEEE}
}

@article{bell1995information,
  title={An information-maximization approach to blind separation and blind deconvolution},
  author={Bell, Anthony J and Sejnowski, Terrence J},
  journal={Neural computation},
  volume={7},
  number={6},
  pages={1129--1159},
  year={1995},
  publisher={MIT Press}
}

@inproceedings{stratos2018mutual,
  title={Mutual Information Maximization for Simple and Accurate Part-Of-Speech Induction},
  author={Stratos, Karl},
  booktitle={Proceedings of the 2019 Conference of the North American Chapter of the Association for Computational Linguistics: Human Language Technologies, Volume 1 (Long and Short Papers)},
  year={2019}
}

@article{sahoo2025simple,
  title={Simple and effective masked diffusion language models},
  author={Sahoo, Subham and Arriola, Marianne and Schiff, Yair and Gokaslan, Aaron and Marroquin, Edgar and Chiu, Justin and Rush, Alexander and Kuleshov, Volodymyr},
  journal={Advances in Neural Information Processing Systems},
  volume={37},
  pages={130136--130184},
  year={2024}
}

@article{cover1978convergent,
  title={A convergent gambling estimate of the entropy of English},
  author={Cover, Thomas and King, Roger},
  journal={IEEE Transactions on Information Theory},
  volume={24},
  number={4},
  pages={413--421},
  year={1978},
  publisher={IEEE}
}

@article{takahira2016entropy,
  title={Entropy rate estimates for natural language—a new extrapolation of compressed large-scale corpora},
  author={Takahira, Ryosuke and Tanaka-Ishii, Kumiko and D{\k{e}}bowski, {\L}ukasz},
  journal={Entropy},
  volume={18},
  number={10},
  pages={364},
  year={2016},
  publisher={MDPI}
}

@article{austin2021structured,
  title={Structured denoising diffusion models in discrete state-spaces},
  author={Austin, Jacob and Johnson, Daniel D and Ho, Jonathan and Tarlow, Daniel and Van Den Berg, Rianne},
  journal={Advances in Neural Information Processing Systems},
  volume={34},
  pages={17981--17993},
  year={2021}
}

@article{campbell2022continuous,
  title={A continuous time framework for discrete denoising models},
  author={Campbell, Andrew and Benton, Joe and De Bortoli, Valentin and Rainforth, Thomas and Deligiannidis, George and Doucet, Arnaud},
  journal={Advances in Neural Information Processing Systems},
  volume={35},
  pages={28266--28279},
  year={2022}
}

@article{onsager1944crystal,
  title={Crystal statistics. I. A two-dimensional model with an order-disorder transition},
  author={Onsager, Lars},
  journal={Physical Review},
  volume={65},
  number={3-4},
  pages={117},
  year={1944},
  publisher={APS}
}

@inproceedings{
butakov2024mutual,
title={Mutual Information Estimation via Normalizing Flows},
author={Ivan Butakov and Alexander Tolmachev and Sofia Malanchuk and Anna Neopryatnaya and Alexey Frolov},
booktitle={The Thirty-eighth Annual Conference on Neural Information Processing Systems},
year={2024},
url={https://openreview.net/forum?id=JiQXsLvDls}
}

@article{newcomb2021use,
  title={Use of Average Mutual Information and Derived Measures to Find Coding Regions},
  author={Newcomb, Garin and Sayood, Khalid},
  journal={Entropy},
  volume={23},
  number={10},
  pages={1324},
  year={2021},
  publisher={MDPI}
}

@inproceedings{darrin2024textttcosmicmutualinformationtaskagnostic,
    title = "{COSMIC}: Mutual Information for Task-Agnostic Summarization Evaluation",
    author = "Darrin, Maxime  and
      Formont, Philippe  and
      Cheung, Jackie  and
      Piantanida, Pablo",
    editor = "Ku, Lun-Wei  and
      Martins, Andre  and
      Srikumar, Vivek",
    booktitle = "Proceedings of the 62nd Annual Meeting of the Association for Computational Linguistics (Volume 1: Long Papers)",
    month = aug,
    year = "2024",
    publisher = "Association for Computational Linguistics",
}

@inproceedings{xiaadanovo,
  title={AdaNovo: Towards Robust$\backslash$emph $\{$De Novo$\}$ Peptide Sequencing in Proteomics against Data Biases},
  author={Xia, Jun and Chen, Shaorong and Zhou, Jingbo and Shan, Xiaojun and Du, Wenjie and Gao, Zhangyang and Tan, Cheng and Hu, Bozhen and Zheng, Jiangbin and Li, Stan Z},
  booktitle={The Thirty-eighth Annual Conference on Neural Information Processing Systems}
}

@book{anderson2012continuous,
  title={Continuous-time Markov chains: An applications-oriented approach},
  author={Anderson, William J},
  year={2012},
  publisher={Springer Science \& Business Media}
}

@article{kelly1981reversibility,
  title={Reversibility and stochastic networks},
  author={Kelly, Frank P},
  journal={PF Kelly—New York: Willy},
  year={1981}
}

@article{macy2024ising,
  title={The Ising model celebrates a century of interdisciplinary contributions},
  author={Macy, Michael W and Szymanski, Boleslaw K and Ho{\l}yst, Janusz A},
  journal={npj Complexity},
  volume={1},
  number={1},
  pages={10},
  year={2024},
  publisher={Nature Publishing Group UK London}
}

@article{cincio2007entropy,
  title={Entropy of entanglement and correlations induced by a quench: Dynamics of a quantum phase transition in the quantum Ising model},
  author={Cincio, Lukasz and Dziarmaga, Jacek and Rams, Marek M and Zurek, Wojciech H},
  journal={Physical Review A—Atomic, Molecular, and Optical Physics},
  volume={75},
  number={5},
  pages={052321},
  year={2007},
  publisher={APS}
}

@article{sherrington1975solvable,
  title={Solvable model of a spin-glass},
  author={Sherrington, David and Kirkpatrick, Scott},
  journal={Physical review letters},
  volume={35},
  number={26},
  pages={1792},
  year={1975},
  publisher={APS}
}

@article{schneidman2006weak,
  title={Weak pairwise correlations imply strongly correlated network states in a neural population},
  author={Schneidman, Elad and Berry, Michael J and Segev, Ronen and Bialek, William},
  journal={Nature},
  volume={440},
  number={7087},
  pages={1007--1012},
  year={2006},
  publisher={Nature Publishing Group UK London}
}

@ARTICLE{2020SciPy-NMeth,
  author  = {Virtanen, Pauli and Gommers, Ralf and Oliphant, Travis E. and
            Haberland, Matt and Reddy, Tyler and Cournapeau, David and
            Burovski, Evgeni and Peterson, Pearu and Weckesser, Warren and
            Bright, Jonathan and {van der Walt}, St{\'e}fan J. and
            Brett, Matthew and Wilson, Joshua and Millman, K. Jarrod and
            Mayorov, Nikolay and Nelson, Andrew R. J. and Jones, Eric and
            Kern, Robert and Larson, Eric and Carey, C J and
            Polat, {\.I}lhan and Feng, Yu and Moore, Eric W. and
            {VanderPlas}, Jake and Laxalde, Denis and Perktold, Josef and
            Cimrman, Robert and Henriksen, Ian and Quintero, E. A. and
            Harris, Charles R. and Archibald, Anne M. and
            Ribeiro, Ant{\^o}nio H. and Pedregosa, Fabian and
            {van Mulbregt}, Paul and {SciPy 1.0 Contributors}},
  title   = {{{SciPy} 1.0: Fundamental Algorithms for Scientific
            Computing in Python}},
  journal = {Nature Methods},
  year    = {2020},
  volume  = {17},
  pages   = {261--272},
  adsurl  = {https://rdcu.be/b08Wh},
  doi     = {10.1038/s41592-019-0686-2},
}

@article{bhanot1988metropolis,
  title={The metropolis algorithm},
  author={Bhanot, Gyan},
  journal={Reports on Progress in Physics},
  volume={51},
  number={3},
  pages={429},
  year={1988},
  publisher={IOP Publishing}
}

@article{pinchas2024comparative,
  title={A Comparative Analysis of Discrete Entropy Estimators for Large-Alphabet Problems},
  author={Pinchas, Assaf and Ben-Gal, Irad and Painsky, Amichai},
  journal={Entropy},
  volume={26},
  number={5},
  pages={369},
  year={2024},
  publisher={MDPI}
}

@article{chai2009exploring,
  title={Exploring functional connectivities of the human brain using multivariate information analysis},
  author={Chai, Barry and Walther, Dirk and Beck, Diane and Fei-Fei, Li},
  journal={Advances in neural information processing systems},
  volume={22},
  year={2009}
}

@inproceedings{nallapati2016abstractive,
    title = "Abstractive Text Summarization using Sequence-to-sequence {RNN}s and Beyond",
    author = "Nallapati, Ramesh  and
      Zhou, Bowen  and
      dos Santos, Cicero  and
      Gu{\ensuremath{\dot{}}}l{\c{c}}ehre, {\c{C}}a{\u{g}}lar  and
      Xiang, Bing",
    editor = "Riezler, Stefan  and
      Goldberg, Yoav",
    booktitle = "Proceedings of the 20th {SIGNLL} Conference on Computational Natural Language Learning",
    month = aug,
    year = "2016",
    publisher = "Association for Computational Linguistics",
    pages = "280--290"
}

@inproceedings{lewis2020bart,
    title = "{BART}: Denoising Sequence-to-Sequence Pre-training for Natural Language Generation, Translation, and Comprehension",
    author = "Lewis, Mike  and
      Liu, Yinhan  and
      Goyal, Naman  and
      Ghazvininejad, Marjan  and
      Mohamed, Abdelrahman  and
      Levy, Omer  and
      Stoyanov, Veselin  and
      Zettlemoyer, Luke",
    editor = "Jurafsky, Dan  and
      Chai, Joyce  and
      Schluter, Natalie  and
      Tetreault, Joel",
    booktitle = "Proceedings of the 58th Annual Meeting of the Association for Computational Linguistics",
    month = jul,
    year = "2020",
    publisher = "Association for Computational Linguistics",
    pages = "7871--7880",
}

@article{libbrecht2015machine,
  title={Machine learning applications in genetics and genomics},
  author={Libbrecht, Maxwell W and Noble, William Stafford},
  journal={Nature Reviews Genetics},
  volume={16},
  number={6},
  pages={321--332},
  year={2015},
  publisher={Nature Publishing Group UK London}
}

@article{whalen2022navigating,
  title={Navigating the pitfalls of applying machine learning in genomics},
  author={Whalen, Sean and Schreiber, Jacob and Noble, William S and Pollard, Katherine S},
  journal={Nature Reviews Genetics},
  volume={23},
  number={3},
  pages={169--181},
  year={2022},
  publisher={Nature Publishing Group UK London}
}

@article{teschendorff2025epigenetic,
  title={Epigenetic ageing clocks: statistical methods and emerging computational challenges},
  author={Teschendorff, Andrew E and Horvath, Steve},
  journal={Nature Reviews Genetics},
  pages={1--19},
  year={2025},
  publisher={Nature Publishing Group UK London}
}

@article{dotan2024effect,
  title={Effect of tokenization on transformers for biological sequences},
  author={Dotan, Edo and Jaschek, Gal and Pupko, Tal and Belinkov, Yonatan},
  journal={Bioinformatics},
  volume={40},
  number={4},
  pages={btae196},
  year={2024},
  publisher={Oxford University Press}
}

@article{qiao2024model,
  title={Model decides how to tokenize: Adaptive dna sequence tokenization with mxdna},
  author={Qiao, Lifeng and Ye, Peng and Ren, Yuchen and Bai, Weiqiang and Liang, Chaoqi and Ma, Xinzhu and Dong, Nanqing and Ouyang, Wanli},
  journal={Advances in Neural Information Processing Systems},
  volume={37},
  pages={66080--66107},
  year={2024}
}

@article{malusare2024understanding,
  title={Understanding the natural language of DNA using encoder--decoder foundation models with byte-level precision},
  author={Malusare, Aditya and Kothandaraman, Harish and Tamboli, Dipesh and Lanman, Nadia A and Aggarwal, Vaneet},
  journal={Bioinformatics Advances},
  volume={4},
  number={1},
  pages={vbae117},
  year={2024},
  publisher={Oxford University Press}
}

@article{eapen2025genomic,
  title={Genomic Tokenizer: Toward a biology-driven tokenization in transformer models for DNA sequences},
  author={Eapen, Bell Raj},
  journal={bioRxiv},
  pages={2025--04},
  year={2025},
  publisher={Cold Spring Harbor Laboratory}
}

@article{grevsova2023genomic,
  title={Genomic benchmarks: a collection of datasets for genomic sequence classification},
  author={Gre{\v{s}}ov{\'a}, Katar{\'\i}na and Martinek, Vlastimil and {\v{C}}ech{\'a}k, David and {\v{S}}ime{\v{c}}ek, Petr and Alexiou, Panagiotis},
  journal={BMC Genomic Data},
  volume={24},
  number={1},
  pages={25},
  year={2023},
  publisher={Springer}
}

@article{elemento2007universal,
  title={A universal framework for regulatory element discovery across all genomes and data types},
  author={Elemento, Olivier and Slonim, Noam and Tavazoie, Saeed},
  journal={Molecular cell},
  volume={28},
  number={2},
  pages={337--350},
  year={2007},
  publisher={Elsevier}
}

@article{rao2007motif,
  title={Motif discovery in tissue-specific regulatory sequences using directed information},
  author={Rao, Arvind and Hero, Alfred O and States, David J and Engel, James Douglas},
  journal={EURASIP Journal on Bioinformatics and Systems Biology},
  volume={2007},
  pages={1--13},
  year={2007},
  publisher={Springer}
}

@article{umarov2017recognition,
  title={Recognition of prokaryotic and eukaryotic promoters using convolutional deep learning neural networks},
  author={Umarov, Ramzan Kh and Solovyev, Victor V},
  journal={PloS one},
  volume={12},
  number={2},
  pages={e0171410},
  year={2017},
  publisher={Public Library of Science San Francisco, CA USA}
}

@article{dreos2013epd,
  title={EPD and EPDnew, high-quality promoter resources in the next-generation sequencing era},
  author={Dreos, Ren{\'e} and Ambrosini, Giovanna and Cavin P{\'e}rier, Rouayda and Bucher, Philipp},
  journal={Nucleic acids research},
  volume={41},
  number={D1},
  pages={D157--D164},
  year={2013},
  publisher={Oxford University Press}
}

@article{hornik1989multilayer,
  title={Multilayer feedforward networks are universal approximators},
  author={Hornik, Kurt and Stinchcombe, Maxwell and White, Halbert},
  journal={Neural networks},
  volume={2},
  number={5},
  pages={359--366},
  year={1989},
  publisher={Elsevier}
}

@article{vaart1997weak,
  title={Weak convergence and empirical processes with applications to statistics},
  author={Vaart, AW van der and Wellner, Jon A},
  journal={Journal of the Royal Statistical Society-Series A Statistics in Society},
  volume={160},
  number={3},
  pages={596--608},
  year={1997},
  publisher={London: Royal Statistical Society, 1988-}
}

@article{chen2025convergence,
  title={Convergence analysis of discrete diffusion model: Exact implementation through uniformization},
  author={Chen, Hongrui and Ying, Lexing},
  journal={Journal of Machine Learning},
  volume={4},
  number={2},
  pages={108--127},
  year={2025}
}

@book{rasmussen2006gaussian,
  title        = {Gaussian Processes for Machine Learning},
  author       = {Rasmussen, Carl Edward and Williams, Christopher K. I.},
  year         = {2006},
  publisher    = {MIT Press},
  address      = {Cambridge, MA},
  isbn         = {0-262-18253-X}
}

@article{scikit-learn,
  title={Scikit-learn: Machine Learning in {P}ython},
  author={Pedregosa, F. and Varoquaux, G. and Gramfort, A. and Michel, V.
          and Thirion, B. and Grisel, O. and Blondel, M. and Prettenhofer, P.
          and Weiss, R. and Dubourg, V. and Vanderplas, J. and Passos, A. and
          Cournapeau, D. and Brucher, M. and Perrot, M. and Duchesnay, E.},
  journal={Journal of Machine Learning Research},
  volume={12},
  pages={2825--2830},
  year={2011}
}

@inproceedings{
zhang2019bertscore,
title={BERTScore: Evaluating Text Generation with BERT},
author={Tianyi Zhang* and Varsha Kishore* and Felix Wu* and Kilian Q. Weinberger and Yoav Artzi},
booktitle={International Conference on Learning Representations},
year={2020},
url={https://openreview.net/forum?id=SkeHuCVFDr}
}

@inproceedings{vedantam2015cider,
  title={Cider: Consensus-based image description evaluation},
  author={Vedantam, Ramakrishna and Lawrence Zitnick, C and Parikh, Devi},
  booktitle={Proceedings of the IEEE conference on computer vision and pattern recognition},
  pages={4566--4575},
  year={2015}
}

@inproceedings{papineni2002bleu,
  title={Bleu: a method for automatic evaluation of machine translation},
  author={Papineni, Kishore and Roukos, Salim and Ward, Todd and Zhu, Wei-Jing},
  booktitle={Proceedings of the 40th annual meeting of the Association for Computational Linguistics},
  pages={311--318},
  year={2002}
}

@INPROCEEDINGS{goebel,
  author={Goebel, B. and Dawy, Z. and Hagenauer, J. and Mueller, J.C.},
  booktitle={IEEE International Conference on Communications, 2005. ICC 2005. 2005}, 
  title={An approximation to the distribution of finite sample size mutual information estimates}, 
  year={2005},
  volume={2},
  number={},
  pages={1102-1106 Vol. 2},
  doi={10.1109/ICC.2005.1494518}}

@article{shi2024simplified,
  title={Simplified and generalized masked diffusion for discrete data},
  author={Shi, Jiaxin and Han, Kehang and Wang, Zhe and Doucet, Arnaud and Titsias, Michalis},
  journal={Advances in neural information processing systems},
  volume={37},
  pages={103131--103167},
  year={2024}
}

@article{nie2025large,
  title={Large language diffusion models},
  author={Nie, Shen and Zhu, Fengqi and You, Zebin and Zhang, Xiaolu and Ou, Jingyang and Hu, Jun and Zhou, Jun and Lin, Yankai and Wen, Ji-Rong and Li, Chongxuan},
  journal={Advances in Neural Information Processing Systems},
  volume={38},
  pages={50608--50646},
  year={2026}
}
